%% file: main.tex
\documentclass[10pt,twocolumn,letterpaper]{article}

\usepackage[pagenumbers]{cvpr} 

\usepackage{amsmath}
\usepackage{multirow}
\usepackage{booktabs}
\usepackage{makecell}
\usepackage{graphicx}
\usepackage{subcaption}
\usepackage{caption}
\usepackage{algorithm}
\usepackage{algpseudocode}
\usepackage{soul}
\usepackage{xcolor}

\definecolor{cvprblue}{rgb}{0.21,0.49,0.74}
\usepackage[pagebackref,breaklinks,colorlinks,allcolors=cvprblue]{hyperref}

\def\paperID{52} 
\def\confName{GRAIL-V}
\def\confYear{2026}

\title{SHOE: Semantic HOI Open-Vocabulary Evaluation Metric}

\author{
{Maja Noack$^{1}$}  \quad
{Qinqian Lei$^{2}$}  \quad
{Taipeng Tian$^{3}$}  \quad
{Bihan Dong$^{1}$} \\
{Robby T. Tan$^{2,4}$}  \quad
{Yixin Chen$^{1}$}  \quad
{John Young$^{1}$}  \quad
{Saijun Zhang$^{1}$}  \quad
{Bo Wang$^1$} \\
$^1$University of Mississippi \;
$^2$National University of Singapore \\
$^3$Independent Researcher \;
$^4$ASUS Intelligent Cloud Services (AICS)\\
\tt\small \url{https://github.com/majnoa/SHOE}
}

\begin{document}

\input{main_paper}

{
    \small
    \bibliographystyle{ieeenat_fullname}
    \bibliography{cvpr2026}
}

\clearpage
\renewcommand{\thesection}{\arabic{section}}
\renewcommand{\thesubsection}{\arabic{section}.\arabic{subsection}}
\renewcommand{\thefigure}{\arabic{figure}}
\renewcommand{\thetable}{\arabic{table}}
\renewcommand{\theequation}{\arabic{equation}}

\setcounter{section}{0}
\setcounter{figure}{0}
\setcounter{table}{0}
\setcounter{equation}{0}

\setcounter{page}{1}

\twocolumn[
\begin{center}
    {\LARGE \bfseries Supplementary Material \par}
    \vspace{2em}
\end{center}
]

\input{supp}

\end{document}

%% file: main_paper.tex
\maketitle

\begin{abstract}
Open-vocabulary human–object interaction (HOI) detection is a step towards building scalable systems that generalize to unseen interactions 
in real-world scenarios and support grounded multimodal systems that reason about human–object relationships.
%
However, standard evaluation metrics, such as mean Average Precision (mAP), treat HOI classes as discrete categorical labels and fail to credit semantically valid but lexically different predictions (e.g., ``lean on couch'' vs. ``sit on couch''), limiting their applicability for evaluating open-vocabulary predictions that go beyond any predefined set of HOI labels.
We introduce SHOE (Semantic HOI Open-Vocabulary Evaluation), a new evaluation framework that incorporates semantic similarity between predicted and ground-truth HOI labels.
SHOE decomposes each HOI prediction into its verb and object components, estimates their semantic similarity using the average of multiple large language models (LLMs), and combines them into a similarity score to evaluate alignment beyond exact string match.
This enables a flexible and scalable evaluation of both existing HOI detection methods and open-ended generative models using standard benchmarks such as HICO-DET.
Experimental results show that SHOE scores align more closely with human judgments than existing metrics, including LLM-based and embedding-based baselines, achieving an agreement of 85.73\% with the average human ratings. 
Our work underscores the need for semantically grounded HOI evaluation that better mirrors human understanding of interactions. 
%
We will release our evaluation metric to the public to facilitate future research.
\end{abstract}

\section{Introduction}
\label{sec:intro}

Human–object interaction (HOI) detection is a fundamental problem in computer vision and artificial intelligence, aiming to understand how humans interact with objects in images. 
Such interaction understanding is also important for grounded multimodal systems, where models must reason about human–object relationships rather than only identifying objects~\cite{xue2025seeing,luo2025thinking,wan2026multimodal}.
Understanding human–object interactions is critical for a wide range of applications, including assistive robotics, autonomous agents, behavioral analysis, and healthcare~\cite{goodrich2008human,suhail2021energy,kong2022human,xie2022chore,xu2023human,liu2026humanmme,ong2026human}.
The combinatorial nature of HOI classes, spanning diverse verbs and objects across varied contexts, makes it infeasible to exhaustively annotate every possible interaction. This intrinsic open-endedness highlights the need for models that generalize beyond fixed label sets, a challenge that is increasingly recognized in HOI research and critical for deploying HOI systems in real-world applications \cite{wu2024dhd,lei2024CMD_SE}.

\begin{figure}[t]
\centering
\includegraphics[width=0.98\columnwidth]{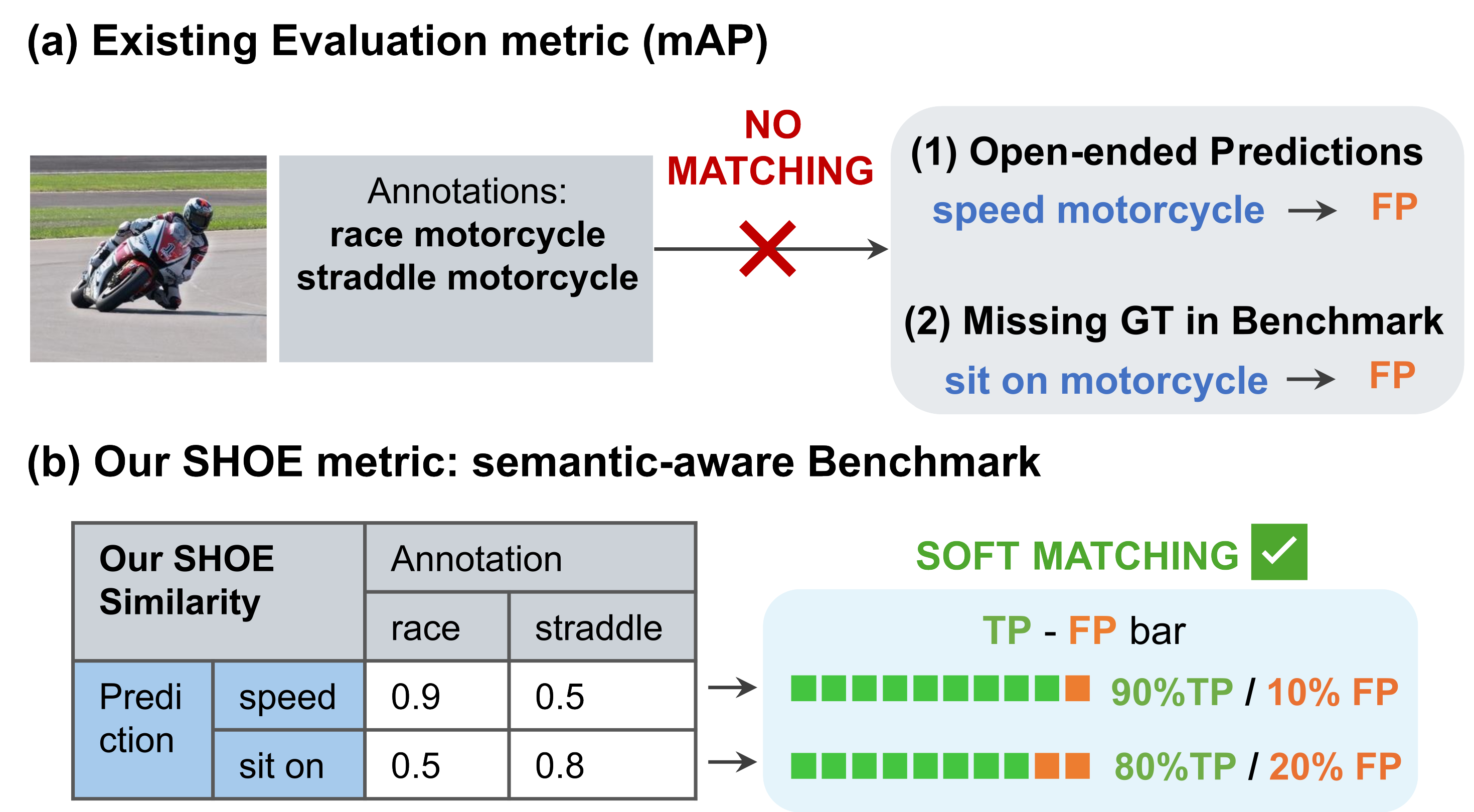}	 
\caption{ (a) Standard mAP metric considers all mismatches as false positives, even when predictions are semantically similar to ground truths. (b) Our SHOE metric assigns soft credit to such cases based on similarity scores, resulting in partial true positives.}
\vspace*{-3mm}
\label{teaser}
\end{figure}

Recent advances in vision-language models (VLMs) and multimodal large language models (MLLMs) have shown strong visual understanding capabilities~\cite{li2022blip,li2023blip2,liu2024LLaVA,bai2025qwen25VL}, presenting a promising avenue for addressing the challenges of open-vocabulary HOI detection.
The standard metric in HOI detection, mean Average Precision (mAP) \cite{chao2018learning}, relies on exact categorical matching, where a prediction is considered correct only if it exactly matches a predefined HOI class in the label set. 
This rigid formulation overlooks the fact that similar or even synonymous interactions, such as 
``race motorcycle'' vs ``speed motorcycle'' are assigned a zero score (see Figure \ref{teaser} a(1)).
%
In open-vocabulary HOI detection, existing exact-match evaluation protocols become fundamentally inadequate, as they undervalue the generalizable outputs of VLMs and MLLMs which are trained on massive datasets instead of limited predefined HOI classes. 
%
%
As a result, existing evaluation frameworks fail to capture the linguistic variability inherent in HOI detection.

In addition to the linguistic variability challenge, existing benchmarks also present challenges for evaluating open-vocabulary predictions.
Datasets such as HICO-DET \cite{hico} and SWIG-HOI \cite{swighoi} 
often suffer from incomplete annotations: not all valid human–object interactions in an image are labeled, which means that reasonable predictions may be penalized as false positives (see Figure \ref{teaser} a(2)). 
For example, the ground truth for an image of a person riding a motorcycle may include only ``race a motorcycle'', while missing equally valid interactions such as ``ride a motorcycle'' or ``straddle a motorcycle''. In this case, a model that predicts ``ride a motorcycle'' or ``straddle a motorcycle'' would be incorrectly penalized.
As a result, semantically valid predictions that differ from the labeled ground truth can be wrongly treated as errors, introducing false negatives and discouraging models from generating diverse or expressive interaction descriptions. 
%
%
These limitations not only undermine the accurate evaluation of open-vocabulary methods but also raise concerns about the reliability of current benchmarks for assessing supervised HOI models, highlighting the need for more flexible and semantically-aware evaluation frameworks. 

To address the limitations of existing evaluation metrics in open-vocabulary HOI detection, we propose SHOE (Semantic HOI Open-Vocabulary Evaluation), a new evaluation framework that supports graded, semantically aware matching.
Instead of requiring exact categorical matches, our approach compares predicted and ground-truth HOI classes by measuring the similarity between their verb and object components. 
Specifically, we compute fine-grained similarity scores using averaged ratings from multiple LLMs, capturing nuanced relationships such as synonymy and relatedness (see Figure \ref{teaser} b).
We do so by assigning each verb and object to its WordNet synset \cite{wordnet}, a sense-specific semantic unit, which removes ambiguity from polysemous words.
These similarity scores for individual HOI predictions are then aggregated across the dataset to produce a final evaluation score.
Importantly, our metric is compatible with both structured HOI class predictions and open-vocabulary outputs, making it broadly applicable to existing HOI detection methods as well as emerging VLMs and MLLMs. 
This enables flexible evaluation across diverse models in open-vocabulary HOI detection.

Our proposed evaluation metric introduces several technical innovations that distinguish it from existing approaches.
First, unlike traditional metrics that rely solely on exact label matches, our method leverages graded similarity scores derived from LLM-based ratings, enabling a more nuanced and semantically aware assessment of model predictions.
Second, we extend our framework to support per-class evaluation, offering insights aligned with existing metrics like mAP.
Third, by decomposing HOI predictions into verb and object components, our metric enables scalable similarity computation, avoiding the combinatorial explosion of direct pairwise comparisons between full verb-object labels, a limitation faced by LLM-derived baselines.
%
%
Together, these components create a flexible and generalizable evaluation protocol that aligns more closely with human judgment and accommodates the growing range of HOI models, including VLMs and MLLMs. Our main contributions are as follows:
\begin{itemize}
    \item A novel evaluation metric for HOI detection that enables graded matching based on semantic similarity, going beyond the binary matches used in existing metrics.
    \item A flexible scoring framework that supports both structured HOI class outputs and class-free open-vocabulary predictions, making it compatible with existing and emerging models, including VLMs and MLLMs.
    \item A scalable decomposition-based design by assessing verb and object similarity separately, avoiding exhaustive HOI pairwise comparisons and supporting significantly larger HOI vocabularies than existing benchmarks.
\end{itemize}
Our user study shows that the proposed metric achieves 85.73\% agreement with the average human rating, surpassing the 78.61\% average agreement among human annotators, highlighting its alignment with human judgment.

\section{Related Work}

\noindent \textbf{HOI Benchmarks}
The field of HOI detection research has been studied across different settings through benchmarks including HICO-DET~\cite{chao2018learning}, V-COCO~\cite{lin2014microsoft} and SWIG-HOI~\cite{wang2022learning}. While recent benchmarks expand the label space up to 10k HOI pairs~\cite{wang2022learning}, they still rely on predefined vocabularies and exact-match evaluation. Even for zero-shot evaluation, HICO-DET is commonly used by holding out a subset of HOI classes during training~\cite{Liao_2022_CVPR}, allowing evaluation on unseen interactions. However, this setting still requires all HOI classes to be predefined prior to inference, and the number of unseen classes is typically limited (e.g., around 120). The current approach limits their flexibility in truly open-world scenarios where models may produce diverse or semantically equivalent interaction descriptions. A concurrent work reformulates HOI detection as a multiple-choice, multiple-answer task to enable evaluation across VLMs and HOI detection methods~\cite{lei2026crosshoi_bench}. Our approach instead focuses on open-vocabulary evaluation, allowing flexible assessment of diverse predictions.

\vspace*{0.1cm}
\noindent \textbf{HOI Detection Methods}
Standard HOI detection methods localize human–object pairs and identify their interactions from a predefined set of HOI classes~\cite{park2023viplo, zhang2021spatially, tamura2021qpic, luo2024discovering}. 
While effective within closed label sets, these methods struggle to generalize to unseen interactions.
To address this limitation, zero-shot HOI approaches leverage pretrained vision-language models (VLMs) like CLIP~\cite{radford2021learning} to tap into external knowledge for novel HOI classes~\cite{ning2023hoiclip, lei2024EZ_HOI, li2024neural, mao2024clip4hoi, cao2024detecting}. 
Open-vocabulary HOI methods go further, aiming to handle a larger and more fine-grained interaction space. CMD‑SE~\cite{lei2024CMD_SE} for example enhances HOI class representations using GPT-generated body-part descriptions.
Despite these advances, state-of-the-art methods still rely on predefined HOI categories during training and inference, limiting their flexibility in open-world scenarios.
Meanwhile, VLMs and MLLMs exhibit the capacity to describe interactions beyond fixed vocabularies. 
However, existing HOI benchmarks and evaluation protocols are not suitable to assess such open-ended predictions, highlighting the need for a new evaluation metric for open-vocabulary HOI detection.

\vspace*{0.1cm}
\noindent \textbf{Evaluation Metrics for HOI}
The standard metric for HOI detection is mean Average Precision (mAP), which evaluates exact matches over a fixed set of HOI categories. While effective in closed-set settings, mAP's exact matching is not suitable for open-vocabulary HOI, where semantically valid predictions may not exactly match labeled ground truth, leading to misclassification.
To address this, several alternative scoring strategies have emerged across vision-language tasks. CLIP-based similarity scoring leverages joint vision-language embeddings to measure the alignment between predicted and reference descriptions.
%
BERT-based similarity~\cite{reimers-gurevych-2019-sentence} and language model plausibility scores~\cite{salazar2020masked} further capture semantic alignment using contextualized textual embeddings or likelihood estimates from pretrained language models.
Incorporating external linguistic resources such as WordNet~\cite{wordnet} also enables structured semantic reasoning. WordNet supports quantifying similarity through distance between sets of cognitive synonyms that share a common meaning, so called synsets, that are structured in directed acyclic graphs (DAGs). 
However, these metrics are based on either lexical similarity or visual similarity, and are not designed to capture HOI similarity, which involves both semantic and visual components.
Building on these insights, we propose a new evaluation metric for HOI detection that incorporates semantic similarity, enabling more robust evaluation in open-vocabulary settings.

\begin{figure*}[t]
\centering
\includegraphics[width=0.95\textwidth]{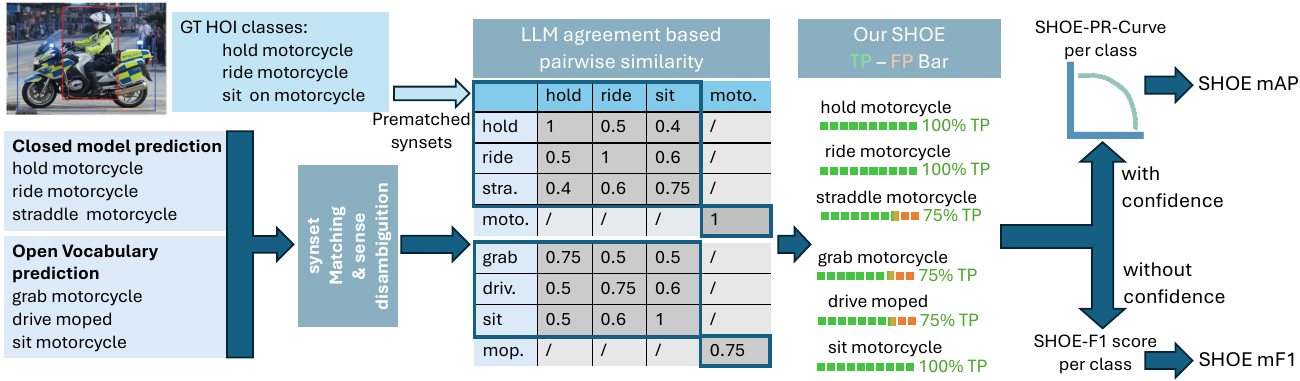} 
\caption{\textbf{SHOE Framework Overview.}
SHOE evaluates closed and open-vocabulary HOI predictions splitting by verb and object mapping to WordNet synsets and computing LLM agreement based pairwise similarity with ground-truth interactions. Predictions get matched to the ground truth based on the highest pair similarity. SHOE mAP is calculated if confidence score is available.
}
\label{methods_fig}
\end{figure*}

\section{Methods}

We propose a semantic similarity-based evaluation framework to address key limitations of existing HOI metrics, such as their reliance on exact label matching and inflexibility to open-vocabulary predictions. 
Our approach reformulates HOI evaluation as a graded matching problem, where predicted and ground-truth HOI classes are compared using semantic similarity scores rather than exact matching. 
%
To support open-vocabulary reasoning, we map verb and object labels from HOI datasets to their corresponding WordNet synsets~\cite{wordnet}. This disambiguation step mitigates lexical ambiguity, ensuring that semantic comparisons reflect intended meanings. 
While this is applicable to any HOI benchmark, we illustrate it on HICO-DET~\cite{hico}, due to its comprehensive and precise annotations. High-annotation completeness is crucial in open-vocabulary evaluation. If valid interactions are missing, semantically correct predictions may be misclassified as false positives.

We then compute fine-grained verb-to-verb and object-to-object similarity scores using averaged ratings from multiple LLMs.
These verb and object similarity scores need to be computed only once per benchmark dataset to build a similarity lookup table and subsequent evaluations simply query the table. 
This makes our framework highly efficient and reusable.
For object labels, we compare against a large neighborhood of synsets in the WordNet hierarchy. For verb labels, due to the shallow and fragmented structure of the verb taxonomy in WordNet, we score each verb synset against a manually curated subset of approximately 7,150 HOI-related synsets (from a total of ~13,000).
Our approach supports both structured HOI class outputs and open-vocabulary verb–object pairs, offering a flexible evaluation protocol that accommodates a wide range of models, including existing HOI detection methods and VLMs or MLLMs. By supporting the full set of verb and object synsets in WordNet (i.e., approximately 126,000 lemmas) our framework enables open-vocabulary evaluation across a wider range of verb–object combinations.


\subsection{Similarity Between Verbs and Objects}
\label{sec:similarity}

Let each predicted Human-Object Interaction (HOI) be represented as $p = (b_h^p, b_o^p, v^p, o^p)$, where $b_h^p$ and $b_o^p$ represent bounding boxes for the human and object, $v^p$ and $o^p$ refer predicted verb and object labels. Similarly, the ground truth for this human-object pair is defined as $g = (b_h^g, b_o^g, v^g, o^g)$. For bounding box matching, we follow the existing evaluation protocol~\cite{Everingham2009,Wang_2019_ICCV}. Please refer to the supplementary material for more details.
%

Next, we define a semantic similarity score between each predicted HOI and each ground-truth HOI that passes the bounding box matching step. Let $\mathcal{M} = \{ (p, g) \mid \mathrm{match}_{\text{box}}(p, g) = 1 \}$ be the set of all predicted ground-truth HOI pairs with matching bounding boxes. For each pair $(p, g) \in \mathcal{M}$, $p = (v^p, o^p)$ and $g = (v^g, o^g)$ are the predicted and ground-truth verb and object labels. We define instance similarity as:
\begin{equation}
    \mathrm{sim}(p, g) = f\left( \mathrm{sim}_v(v^p, v^g), \mathrm{sim}_o(o^p, o^g) \right),
\end{equation}
\noindent where $\mathrm{sim}_v(v^p, v^g) \in [0, 1]$ and $\mathrm{sim}_o(o^p, o^g) \in [0, 1]$ denote the verb and object similarity scores, respectively. The function $f(\cdot, \cdot)$ combines these scores into an overall instance-level similarity. 
%
We explore several aggregation strategies, including weighted mean, geometric mean ($f_{\text{geom}}(v,o) = \sqrt{vo}$), and minimum ($f_{\min}(v,o) = \min(v,\, o)$). In our model evaluation, we use the standard arithmetic mean, $f_\text{arith}(v,o) = w\,v + (1-w)$, with $w=\frac{1}{2}$.

\paragraph{Instance-wise Matching Score}
We now define a soft matching score per individual ground-truth HOI, based on the most similar prediction. This allows us to compute soft true positives, false positives, and false negatives at the instance level while accounting for semantic similarity. Let $\mathcal{G} = \{ g_1, g_2, \dots, g_m \}$ be the set of all ground-truth HOIs, and $P = \{ p_1, p_2, \dots, p_n \}$ be the set of predicted HOIs. For each ground-truth HOI $g_j \in \mathcal{G}$ , we compute an instance-wise matching score $t_j$,
\begin{equation}
\begin{aligned}
    & t_j = \max_{p \in P} \mathrm{sim}(p, g_j), \\
    & p_{g_j} = \arg \max_{p \in P} \mathrm{sim}(p, g_j)
\end{aligned}
\end{equation}
where $t_j \in [0, 1]$ reflects how well the prediction set $P$ covers ground-truth instance $g_j$  
and $p_{g_j}$ denotes the matched prediction for $g_j$. We collect all such matched predictions into a set $\mathcal{P}_G = \{ p_{g_j} \mid g_j \in \mathcal{G} \}$.

{
Let $\mathcal{C} = \{c_1, c_2, \dots, c_k\}$ be the set of all unique GT HOI classes in the dataset, where $c = (v, o)$.  For each predicted HOI $p_i \in P$, we compute a prediction matching score $s_i$,
\begin{equation}
\begin{aligned}
    & s_i = \max_{g \in \mathcal{C}} \mathrm{sim}(p_i, g), \\
    & c_{s_i} = \arg \max_{g \in \mathcal{C}} \mathrm{sim}(p_i, g)
\end{aligned}
\end{equation}
where $s_i$ measures similarity between the predicted $p_i$ and its most similar ground-truth class $c_{s_i}$ in the evaluation set.}

\subsection{Dataset-level Aggregation}
Standard evaluation metrics such as mAP rely on confidence scores to rank predictions. 
For VLMs, however, no native HOI confidence score exists. Recent studies have shown that prompted confidence scores from VLMs and MLLMs may not reliably reflect true uncertainty, especially in open-ended tasks~\cite{zhou2024relying, chhikara2025mind}. 
Alternatively the token probability can be used as a proxy for an HOI confidence score but it reflects only models token-generation dynamics rather than the correctness of the predicted HOI.

To account for this, we provide two evaluation modes: category (1) a confidence-based mode that aligns with mAP-style ranked evaluation (token probabilities are used for VLMs), and category (2) a confidence-free mode that evaluates all predictions equally. 
This design allows for both compatibility with conventional benchmarks and robustness to potential miscalibration.
Figure \ref{methods_fig} shows the workflow of the two aggregation strategies.
We begin by computing soft precision, recall, and F1 scores per HOI class, based on the instance-wise matching score. This allows evaluation that aligns with standard mAP-style metrics while supporting semantic soft matching.
The confidence score rank is denoted with $i$. Here $i$ denotes the prefix of the ranked prediction list up to position i if confidence scores are available.
If no confidence scores are used, $i$ is dropped and the calculation is done using all predictions of the class. 
$G_c$ be the set of ground-truth instances of class $c \in \mathcal{C}$.
For each class $c \in \mathcal{C}$, we define True Positives (TP) as:
\begin{equation}
\text{TP}_c^{(i)} = \sum_{g_j \in G_c^{(i)}} t_j
\end{equation}
False Negatives (FN):
\begin{equation}
\text{FN}_c^{(i)} = \sum_{g_j \in G_c^{(i)}} (1 - t_j)
\end{equation}
False Positives (FP):
\begin{equation}
\text{FP}_c^{(i)} = \sum_{p_{k} \in \mathcal{P}_G^{(i)}} 1- t_k + \sum_{p_{k} \notin \mathcal{P}_G^{(i)}} 1
\end{equation} 
The first term captures predictions matched to ground-truth instances within $G_c^{(i)}$, but not exact matching ($t_j \neq 1$). The second term includes unmatched predictions. 


\noindent Soft Precision:
\begin{equation}
\text{Precision}_c^{(i)} = \frac{\text{TP}_c^{(i)}}{\text{TP}_c^{(i)} + \text{FP}_c^{(i)}}
\end{equation}
Soft Recall:
\begin{equation}
\text{Recall}_c^{(i)} = \frac{\text{TP}_c^{(i)}}{\text{TP}_c^{(i)} + \text{FN}_c^{(i)}}
\end{equation}

The F1-Score can be calculated for every model. However, for models of category (1) a set
confidence score rank $i$ needs to be chosen to allow for fair comparison. The Soft F1 Score is defined as:

\begin{equation}
\text{F1}_c^{(i)} = \frac{2 \cdot \text{Precision}_c^{(i)} \cdot \text{Recall}_c^{(i)}}
{\text{Precision}_c^{(i)} + \text{Recall}_c^{(i)}}
\end{equation}

These scores can be aggregated into the final mean soft F1 score over all GT HOI classes. Mean Soft F1 Score:

\begin{equation}
    \text{mF1} = \frac{1}{|\mathcal{C}|} \sum_{c \in \mathcal{C}} \text{F1}_c^{(i)}
\end{equation}

Additionally for models of category (2) the Soft Average Precision can be calculated from the Area under the Precision-Recall-Curve. Soft Average Precision (Soft-AP):
\begin{equation}
    \text{Soft-AP}_c = \sum{i} \left( \text{Recall}_c^{(i)} - \text{Recall}_c^{(i-1)} \right) \cdot \text{Precision}_c^{(i)}
\end{equation}
Similar to the F1 score these Soft-APs can be aggregated into the mean over all classes. Soft mean Average Precision (Soft-mAP) is defined as: 
\begin{equation}
\text{Soft-mAP} = \frac{1}{|\mathcal{C}|} \sum_{c \in \mathcal{C}} \text{Soft-AP}_c
\end{equation}

\section{Experiments}

\subsection{Experiment Setting}

\noindent \textbf{Evaluation Dataset}
We build our evaluation on top of the widely used HICO-DET dataset~\cite{hico}, which contains 600 predefined HOI classes, each defined by a verb–object pair. 
While HICO-DET provides a foundation for HOI detection, its label space is limited to a small and fixed set of 117 verbs and 80 objects.
To support open-vocabulary evaluation, we map each verb and object in HICO-DET to its corresponding WordNet synset. This allows us to expand the HOI label space by including semantically related concepts in the WordNet hierarchy (e.g., hypernyms, hyponyms, and sibling nodes). For example, the verb “hold” in HICO-DET can be connected to related verbs such as “clench," “cling," or “grasp." 
Similarly, the object “bicycle” can be linked to related concepts like “vehicle” or “mountain bike.”
This WordNet-based vocabulary expansion produces a significantly broader set of candidate HOI classes while maintaining semantic alignment with the original dataset.

\vspace*{0.1cm}
\noindent \textbf{Evaluation Metrics}
The standard evaluation metric for human–object interaction (HOI) detection is mAP, which assesses performance based on spatial alignment of predicted and ground-truth  human and object bounding boxes and exact matching of verb and object class labels. 
This metric is used across benchmark datasets such as HICO-DET \cite{hico} and SWIG-HOI \cite{swighoi}.
In this work, we proposed a new semantically aware evaluation metric SHOE that relaxes the binary matching constraint in mAP by introducing graded similarity between predicted and ground-truth HOI classes, resulting in a soft mAP score or SHOE-score that captures partial semantic matches.

\vspace*{0.1cm}
\noindent \textbf{Implementation Details}
To compute the pairwise verb–verb and object–object similarities that underlie our metric, we begin by mapping the verbs and objects in HICO-DET to their corresponding synsets in WordNet. 
Using the WordNet graph structure, we expand each label to include semantically related nodes, such as parent and child synsets for objects, and a manually annotated set of 7,150 HOI verb candidates, resulting in a significantly broader HOI label vocabulary.
We then use LLMs to compute similarity scores between all candidates and ground truth verbs and objects in this expanded label space. 
Each LLM rated the similarity between labels on a five-point scale,  
from completely dissimilar (0) to synonymous (4), using the glosses of the corresponding synsets as reference.

\begin{figure}[t]
    \centering
    \includegraphics[width=0.85\linewidth]{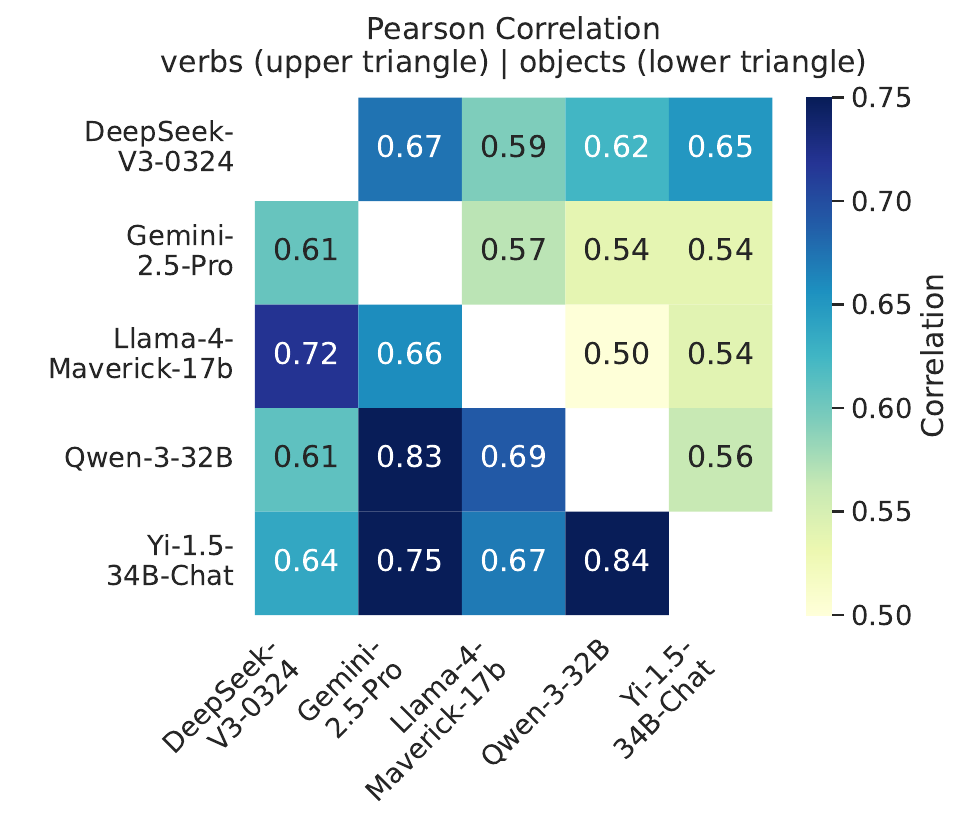} 
    \caption{
    Pearson correlation between LLMs for verb (upper triangle) and object (lower triangle) similarity ratings.}
    \label{fig:llm_pearson}
\end{figure}

To handle the computational load of exhaustively comparing ~7k candidate verbs against 117 GT verbs from HICO-DET ($\approx 850\text{K}$ comparisons), we use Qwen3-32B~\cite{yang2025qwen3} to perform initial exhaustive pairwise comparisons. 
This yields a sparse similarity matrix, with most scores being zero due to semantic dissimilarity. 
For object pairs and the subset of verb pairs with non-zero similarity, we collect additional judgments using four more LLMs, DeepSeek-V3-0324~\cite{liu2024deepseek}, Llama-4-Maverick-17B~\cite{touvron2023llama}, Yi-1.5-34B-Chat~\cite{young2024yi}, and Gemini-2.5-pro~\cite{comanici2025gemini}, to improve robustness through cross-model agreement.
We average these model outputs to produce similarity scores that serve as the basis for evaluating semantic alignment between predicted and ground truth HOI labels.
To assess consistency across models, we compute Pearson correlation between their similarity scores (see Figure~\ref{fig:llm_pearson}), which quantifies the degree of linear score alignment between pairs of models.

We find that the object similarity ratings of LLMs (lower triangle) demonstrate better agreement between models than the verb similarity ratings (upper triangle).
The strongest object correlation occurs between Qwen3-32B and Yi-1.5-34B-Chat ($r = 0.84$), while verb correlations vary more widely, ranging from 0.50 to 0.72, highlighting how the broader semantic range and polysemy of actions complicate verb similarity scoring.

\subsection{Validation of Pairwise Verb Synset Filtering}
To verify that using Qwen3-32B to filter verb synset pairs did not unintentionally exclude semantically meaningful verb pairs, we conducted a control experiment. We randomly sampled 100k verb synset pairs from the pool of excluded pairs (Qwen3-32B similarity rating = 0) and re-evaluated them using the other LLMs included in SHOE. Across models, disagreement with Qwen’s zero-similarity judgments remained minimal: 1.318 \% Llama 4 Maverick-17B, 0.903 \% DeepSeek-V3, 1.105 \% Yi-1.5-34B-Chat, and 0.245 \% Gemini 2.5 Pro. This shows that the filtering step preserves relevant similarity coverage while substantially reducing computational cost.

\subsection{User Study}

We conducted a user study with five annotators to evaluate how well our HOI similarity metric reflects human judgments.
The study involved 500 HOI pair comparisons, each consisting of a ground-truth interaction from the HICO-DET dataset and an artificially generated prediction. The ground-truth pairs were selected to reflect the class distribution in HICO-DET.
Using our SHOE similarity metric, we generated predicted HOI interactions that spanned diverse semantic relationships, grouped into three categories: (1) same verb, different object; (2) different verb, same object; and (3) different verb and different object. For each category, we sampled predicted HOI pairs at four distinct similarity levels, based on their SHOE scores (ranging from 1 to 4), with an approximately equal number of examples per level.
Annotators were shown each HOI pair along with WordNet glosses for all involved synsets and were asked to rate the similarity between the two on a 5-point scale: 0 (completely dissimilar), 1 (somewhat related but clearly distinct), 2 (related but easily distinguishable), 3 (very similar with subtle differences), and 4 (interchangeable).

\begin{figure}[t]
    \centering
    \includegraphics[width=0.7\linewidth]{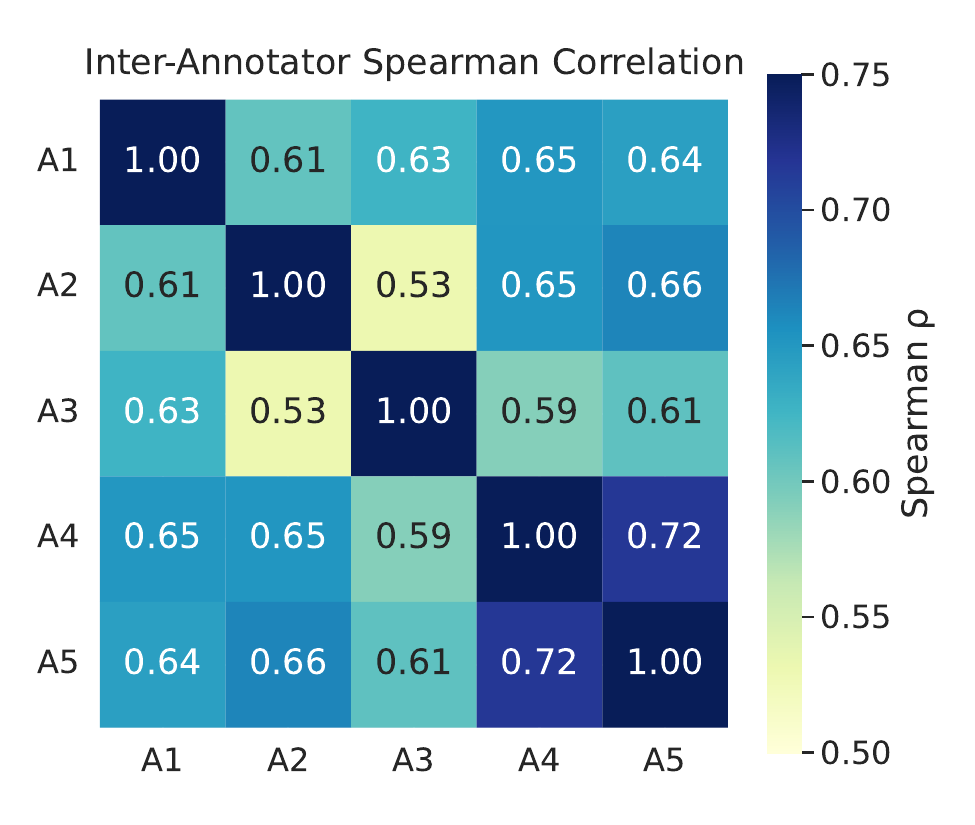}
    \caption{
    Pairwise Spearman rank correlation ($\rho$) between annotators (A1–A5) on HOI similarity ratings. Higher values indicate stronger alignment in their relative rankings of interaction pairs.
    }
    \label{fig:userstudyheatmap}
    
\end{figure}

%
Our study shows moderate to strong agreement between annotators in Figure~\ref{fig:userstudyheatmap}. The Spearman’s $\rho$ values ranged between 0.531 and 0.716, with most values above 0.6, a threshold commonly considered indicative of moderate to strong agreement \cite{Akoglu2018}. 
Additionally we compute the average agreement score between annotators. 
For each pair of annotators, we compute agreement scores over the set of items they both rated. Given a ground-truth rating $r_i$ and a comparison rating $r_j$ for the same item, the agreement score is defined as:
\begin{equation}
\label{eq:pairagree}
\text{Agreement}(r_i, r_j) = 1 - \frac{|r_i - r_j|}{4}
\end{equation}
The pairwise agreement is then calculated as the mean of these scores over all shared items. Finally, the overall agreement score is computed as the average of all pairwise agreement values across annotator pairs.
The average agreement among annotators is 78.61\%.

\begin{figure}[t]
    \centering
   \includegraphics[width=0.95\linewidth]{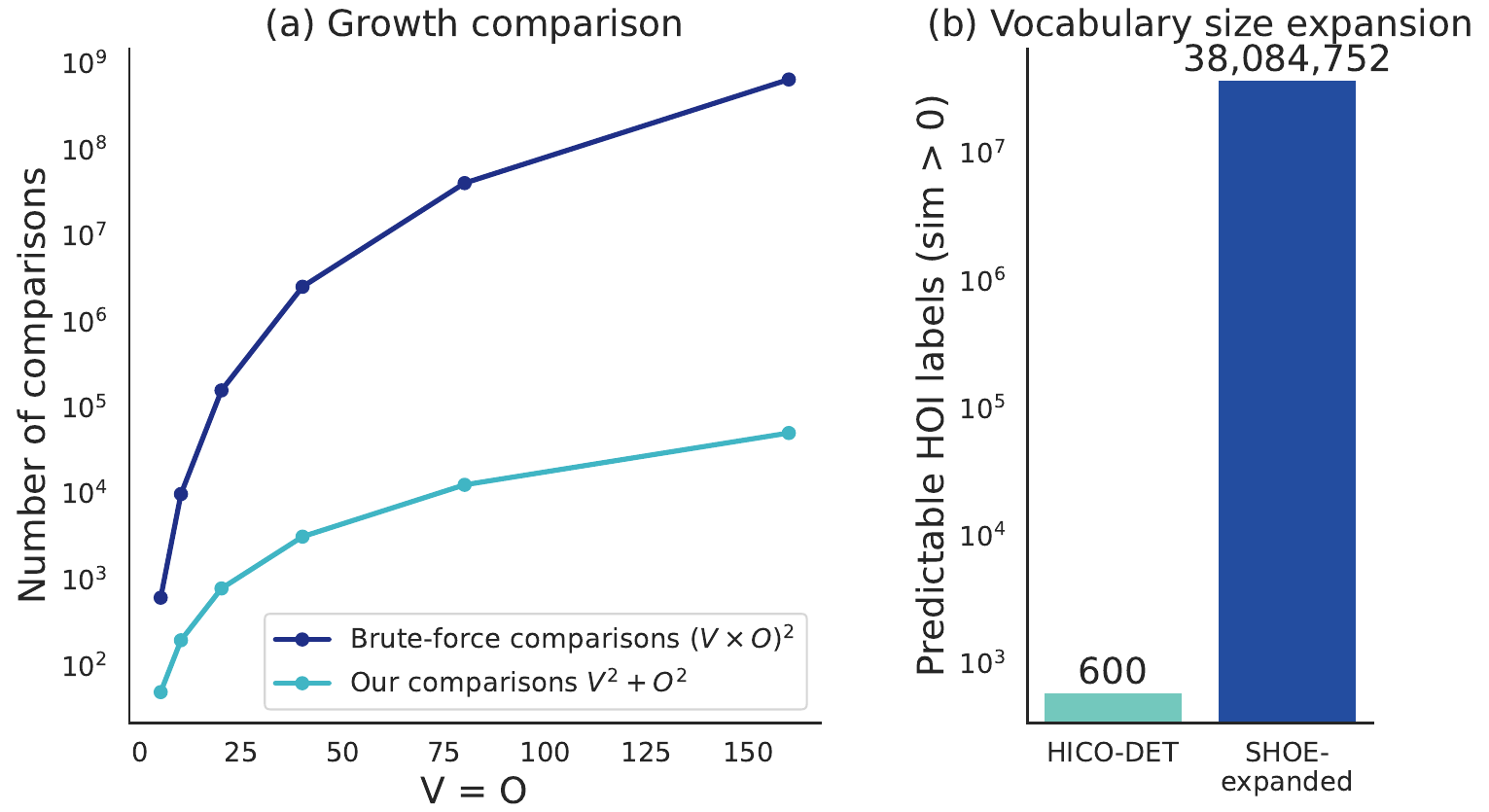}
    \caption{
    (a) Comparison of the number of pairwise similarity computations required by brute-force HOI evaluation versus our approach. %
    The brute-force method grows as $(V \times O)^2$, while our method grows as $V^2 + O^2$. (b) SHOE expands the 600 HOI classes from HICO-DET to over 38M semantically related HOI labels.
    }
    \label{fig:hoi-growth-comparison}
\end{figure}

\subsection{Evaluation of Similarity Metrics}
Based on our user study, we evaluated how closely different HOI similarity metrics align with human judgment by comparing their scores to average human ratings, as summarized in Table~\ref{tab:userstudyagreement}.
The average agreement is computed using the same formula defined in Equation~\ref{eq:pairagree}.
Our proposed SHOE metric, using the standard arithmetic mean, achieved the highest agreement with annotators (85.73\%), surpassing the average agreement among human annotators (78.61\%).
Alternative aggregation strategies, such as the geometric mean or taking the minimum verb or object SHOE score, yield slightly lower agreement but remain competitive.
In contrast, the instance-based direct-matching score used in mAP has the lowest agreement with human ratings (38.90\%), confirming that rigid exact matching is insufficient for open-vocabulary HOI evaluation.

We also evaluated a set of LLMs, including DeepSeek-V3-0324~\cite{liu2024deepseek}, Llama-4-Maverick-17B~\cite{touvron2023llama}, Yi-1.5-34B-Chat~\cite{young2024yi}, and Gemini-2.5-pro~\cite{comanici2025gemini}, by prompting them to directly rate the similarity between full HOI pairs.
These models show high alignment with human judgments, with agreement scores ranging from 77.52\% to 83.34\%. 
By contrast, our method, decomposing HOI similarity into independent verb and object similarities, achieves better alignment and is significantly more scalable. 
Instead of scoring every possible verb-object pair combination (which grows quadratically with the number of HOI classes), we only need to compute pairwise similarities among verbs and among objects (see Figure \ref{fig:hoi-growth-comparison} a). 
%
This design drastically increases the HOI vocabulary, expanding the 600 discrete HICO-DET labels into over 38M semantically related HOIs (see Figure \ref{fig:hoi-growth-comparison} b) while maintaining computational efficiency.
These similarity scores have to be computed only a single time per dataset to build a similarity lookup table, and subsequent evaluations simply query the table.

CLIP is known for its strength in aligning vision and language, however, it performs poorly in HOI pair similarity comparisons, achieving only 59.11\% agreement with human annotations.
Classical WordNet-based graph distance metrics, such as Wu–Palmer (WUP), Leacock–Chodorow (LCH), and Path similarity, also perform poorly, with agreement scores ranging from 52\% to 57\%. 
This suggests that although WordNet’s taxonomy encodes human-understandable lexical relationships, its hierarchical distances are insufficient to capture the nuanced semantics required for HOI similarity.
Similarly, sentence embedding models such as SentenceBERT ~\cite{Reimers2019} and BGE ~\cite{Zhao2023} yield comparably low agreement scores. These models, while effective for general semantic similarity tasks, appear to lack the fine-grained compositional understanding needed to distinguish subtle variations in human-object interactions.

\begin{table}[t]
\centering
\small
\renewcommand{\arraystretch}{1.15}
\setlength{\tabcolsep}{3pt}

\begin{tabular}{l l c}
\toprule
\textbf{Category} & \textbf{Metric} & \textbf{Agreement (\%)} \\
\midrule

\multirow{3}{*}{\textbf{SHOE}} 
  & Standard & \textbf{85.73} \\
  & Geometric Mean & 84.29 \\
  & Minimum & 84.01 \\
\midrule

\multirow{6}{*}{\textbf{LLMs}}
  & DeepSeek-V3-0324 & 83.34 \\
  & Qwen3-32B & 82.03 \\
  & Llama-4-Mav-17B & 82.10 \\
  & Yi-1.5-34B-Chat & 81.55 \\
  & Gemini-2.5-Pro & 77.52 \\
\midrule

\multirow{1}{*}{\textbf{WordNet}}
  & WUP / LCH / Path & 57.09 / 55.69 / 52.47 \\
\midrule

\multirow{2}{*}{\textbf{Embeddings}}
  & CLIP-ViT-B (gloss/pair) & 59.11 / 58.15 \\
  & BGE (gloss/pair) & 57.41 / 56.37 \\
  & SBERT (gloss/pair) & 54.09 / 52.86 \\
\midrule

\multirow{1}{*}{\textbf{mAP}}
  & direct-match & 38.90  \\
\bottomrule
\end{tabular}

\caption{
Agreement between HOI similarity metrics and average human judgments collected via a user study. Each metric scores the similarity of human-object interaction (HOI) pairs. For methods that compute cosine similarity, they either use embeddings of HOI phrases (pair) or embeddings of their WordNet glosses (gloss). WordNet-based metrics are semantic graph distances between verb-object synset pairs. 
}
\label{tab:userstudyagreement}
\end{table}

\subsection{Hyperparameter Tuning}
\begin{figure}[t]
    \centering
    \includegraphics[width=0.9\linewidth]{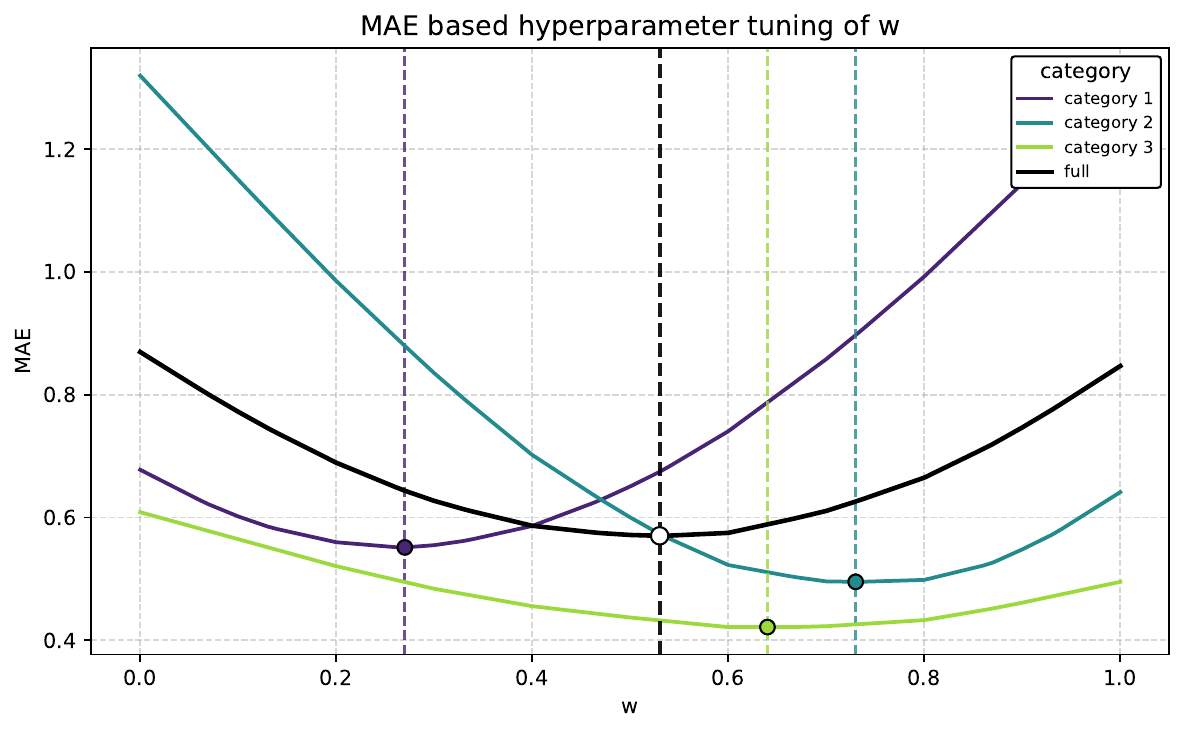}
    \caption{
    Optimizing the arithmetic-mean weight $w$ via MAE for each of the three categories in our user study. Category 1 (same verb, different object), Category 2 (different verb, same object), Category 3 (both different) 
    }
    \label{fig:hyperpram_w}
\end{figure}

To determine how strongly verb similarity versus object similarity should contribute to the final HOI similarity score, we conduct a hyperparameter ablation minimizing the Mean Absolute Error (MAE) over the weighting factor w in the arithmetic mean (see Section~\ref{sec:similarity}).
Let each example belong to a semantic category $c(i)\in\{1,\dots,C\}$, and let
$v_i$, $o_i$ and $y_i$ denote the verb and object similarities and human-annotated similarity score associated with prediction $i$.
For each category $c$, we compute the MAE of a weighted arithmetic mean
$w v_i + (1-w)o_i$ as
\begin{equation}
\mathrm{MAE}_c(w)
    = \frac{1}{N_c}
      \sum_{i:\,c(i)=c}
      \bigl|\, w\, v_i + (1-w)\, o_i - y_i \,\bigr| .
\end{equation}
The optimal weight $w_c^\star$ for category $c$ is the value of $w$ that minimizes the MAE. 
The results of the hyperparameter tuning are shown in Figure \ref{fig:hyperpram_w}. We observe that Category 1 (same verb, different object) favors the object term, yielding a low optimum of $w_1^\star = 0.267$; 
Category 2 (different verb, same object) relies more heavily on the verb term, with $w_2^\star = 0.733$; and Category 3 (both different) slightly emphasizes verbs, giving $w_3^\star = 0.640$. The overall agreement increases to 87.63\% by category based tuning. 
As the user study is relatively small in size ($\sim$166 pairs per category), we set $w=0.5$ in all experiments.

\subsection{Evaluated Models}
We compare our proposed SHOE mAP with the standard mAP on HICO-DET across four settings: default, zero-shot rare first unseen combination (RF-UC), and open-vocabulary and VLMs with DETR \cite{carion2020end} aided object detection (see Table~\ref{tab:hoi_results_closed}).
The models span recent state-of-the-art HOI detectors, including LAIN~\cite{kim2025locality}, CMMP~\cite{lei2024exploring}, ADA-CM~\cite{lei2023efficient}, and HOLA~\cite{lei2025hola}, EZ-HOI~\cite{lei2024EZ_HOI} under default and zero-shot configurations, as well as CMD-SE~\cite{lei2024CMD_SE} and THID~\cite{wang2022learning} for open-vocabulary generalization and state-of-the-art VLMs like Qwen2.5-32B-VL \cite{bai2025qwen25VL}, InternVL3-38B \cite{chen2024internvl}, and GPT-4.1 \cite{2025Achiamgpt}. 
Existing VLMs exhibit limited grounding capabilities and are unable to predict bounding boxes reliably. 
Therefore, we first predict human and object bounding boxes using DETR. In the model prompt we provide the DETR detections and task the models to predict the interaction verb. We use the verb token probabilities as a proxy for confidence score.

As shown in Table~\ref{tab:hoi_results_closed}, SHOE mAP produces slightly higher scores than standard mAP in the default and zero-shot settings by rewarding semantically related predictions that differ from the ground truth, thus reducing the false negative rate and better reflecting human understanding of interaction similarity.
HOLA (ViT-L) achieves the highest SHOE-score with 39.92 in the default setting, while LAIN (ViT-B) reaches a 35.37 score in the zero-shot setting.
The open-vocabulary models show lower overall mAP and SHOE mAP scores among all evaluated methods.
THID reaches the highest SHOE mAP in this category with 22.04.

The results of VLMs with DETR highlight the limitations of strictly instance-based metrics like mAP. All models achieve significantly higher SHOE mAP than standard mAP. 
GPT-4.1 exhibits the highest standard mAP (49.50) outperforming other VLMs. Without being trained on the dataset, it most accurately reproduces the exact HOI labels defined in HICO-DET. Qwen2.5-VL-32B has the lowest standard mAP among the three VLMs showing lower exact-match accuracy. 
However, it achieves the highest SHOE-mAP (66.03) of all tested models, indicating strong semantic and visual understanding of HOIs. This confirms that SHOE reveals meaningful distinctions that mAP fails to capture and underlines the HOI capabilities of VLMs.
SHOE mAP offers a more nuanced evaluation by incorporating semantic similarity beyond exact verb-object pair matching, while still preserving compatibility with standard mAP scoring for exact matches.

\begin{table}[t]
\centering
\small 
\renewcommand{\arraystretch}{1.1} 
\setlength{\tabcolsep}{4.5pt} 
\begin{tabular}{llcc}
\toprule
  \makecell{\textbf{Method}\\\textbf{Type}} & \textbf{Model}  & \textbf{mAP} & \makecell{\textbf{SHOE}\\\textbf{mAP}} \\
\midrule

\multirow{4}{*}{\textbf{Default}} 
  & LAIN (ViT-B)        &  35.30 & 36.58 \\
  & CMMP (ViT-L)        &  37.77 & 39.06 \\
  & ADA-CM (ViT-L)      &  38.40 & 39.72 \\
  & HOLA (ViT-L)        &  \textbf{39.05} & \textbf{39.92} \\
\midrule
\multirow{4}{*}{\makecell{\textbf{Zero-shot}\\ \textbf{RF-UC}}} 
  & CMMP (ViT-B)     & 32.18 &  32.95\\
  & EZ-HOI (ViT-B)   &  33.13 &  33.83\\
  & HOLA (ViT-B)     &  34.19 &  35.26\\
  & LAIN (ViT-B)     &  \textbf{34.60} &  \textbf{35.37}\\
\midrule
\multirow{2}{*}{\makecell{\textbf{Open-}\\\textbf{Vocabulary}}} 
  & CMD-SE           &  18.20  & 19.03 \\
  & THID              &  \textbf{22.01} & \textbf{22.04} \\
\midrule
\multirow{3}{*}{\textbf{DETR + VLMs}} 
  & GPT-4.1       &  \textbf{49.50} & 61.67 \\
  & InternVL3-38B  & 42.00 & 58.03\\
  & Qwen2.5-VL-32B    &  34.83 & \textbf{66.03}\\  
\bottomrule
\end{tabular}
\caption{Performance comparison across different HOI models and evaluation metrics on HICO-DET, including our proposed SHOE mAP and standard mAP. For VLMs, token probabilities are used as a proxy for the confidence score.} 
\label{tab:hoi_results_closed}
\vspace*{-3mm}
\end{table}

\section{Conclusion}
We introduced SHOE, a new evaluation metric for HOI detection that captures semantic similarity between verb-object pairs by decomposing interactions into their verb and object components. 
By leveraging LLMs to compute pairwise similarities, SHOE enables a fine-grained assessment of model predictions, especially under open-vocabulary condition.
Our user study demonstrates that SHOE aligns more closely with human judgment than existing metrics. 
%
The modular structure of SHOE also allows for efficient scaling with larger HOI vocabulary. 
We hope SHOE will serve as a practical evaluation metric for future HOI models, going beyond the limits of exact class matching.

%% file: supp.tex
\begin{table*}[t]
\centering
\small
\begin{minipage}{0.45\linewidth}
\centering
\begin{tabular}{ccccccc}
\toprule
\textbf{\#} & DS & G. & Mv. & Qw. & Yi & Agree.\% \\
\midrule
1 &1&0&0&0&0 & 86.46 \\
1 &0&1&0&0&0 & 84.56 \\
1 &0&0&1&0&0 & 84.11 \\
1 &0&0&0&1&0 & 85.60 \\
1 &0&0&0&0&1 & 84.87 \\
Avg. &&&&& & 85.12$\pm$0.92 \\
\midrule
2 &1&1&0&0&0 & 86.46 \\
2 &1&0&1&0&0 & 86.46 \\
2 &1&0&0&1&0 & 86.46 \\
2 &1&0&0&0&1 & 86.46 \\
2 &0&1&1&0&0 & 84.56 \\
2 &0&1&0&1&0 & 84.56 \\
2 &0&1&0&0&1 & 84.56 \\
2 &0&0&1&1&0 & 84.58 \\
2 &0&0&1&0&1 & 84.15 \\
2 &0&0&0&1&1 & 85.60 \\
Avg. &&&&& & 85.39$\pm$0.99 \\
\bottomrule
\end{tabular}
\end{minipage}
\hfill
\begin{minipage}{0.45\linewidth}
\centering
\begin{tabular}{ccccccc}
\toprule
\textbf{\#} & DS & G. & Mv. & Qw. & Yi & Agree.\% \\
\midrule
3 &1&1&1&0&0 & 85.55 \\
3 &1&1&0&1&0 & 85.76 \\
3 &1&1&0&0&1 & 86.12 \\
3 &1&0&1&1&0 & 85.40 \\
3 &1&0&1&0&1 & 85.52 \\
3 &1&0&0&1&1 & 85.93 \\
3 &0&1&1&1&0 & 84.71 \\
3 &0&1&1&0&1 & 84.53 \\
3 &0&1&0&1&1 & 84.88 \\
3 &0&0&1&1&1 & 85.12 \\
Avg. &&&&& & 85.35$\pm$0.53 \\
\midrule
4 &1&1&1&1&0 & 85.67 \\
4 &1&1&1&0&1 & 85.86 \\
4 &1&1&0&1&1 & 86.40 \\
4 &1&0&1&1&1 & 85.91 \\
4 &0&1&1&1&1 & 85.24 \\
Avg. &&&&& & 85.82$\pm$0.42 \\
\midrule
5 (SHOE) &1&1&1&1&1 & 85.73 \\
\bottomrule
\end{tabular}
\end{minipage}

\vspace{2mm}

\caption{
Combined LLM ablation: left = sizes 1-2, right = sizes 3-5. Agreement scores for LLMs with human annotation. 1 = model included, 0 = excluded. \# = number of models included.
LLMs in order: DeepSeek-V3-0324, Gemini-2.5-pro, Llama-4-Maverick-17B, Qwen3-32B, Yi-1.5-34B-Chat.
}
\label{tab:llm_ablation_combined}
\end{table*}

\section{LLM Ablation for SHOE-Metric}
To assess the sensitivity of SHOE to the specific set of LLMs (Qwen3-32B~\cite{yang2025qwen3}, DeepSeek-V3-0324~\cite{liu2024deepseek}, Llama-4-Maverick-17B~\cite{touvron2023llama}, Yi-1.5-34B-Chat~\cite{young2024yi}, and Gemini-2.5-pro~\cite{comanici2025gemini}) used to compute the final similarity score and whether a smaller or differently composed ensemble could achieve comparable alignment with human judgments, we evaluated all possible subsets of the five LLMs used in our ensemble, ranging from single-model configurations to the full five-model setup (Table \ref{tab:llm_ablation_combined}).
Across all configurations, agreement stays within a narrow 0.84–0.86 range. Single models already achieve strong alignment, though performance varies (e.g., DeepSeek-V3 is highest at 86.46 \%; Maverick lowest at 84.11 \%). Adding a second model provides only marginal benefit: most two-model combinations match the performance of the stronger individual model, indicating limited complementary information at this scale. Larger ensembles (3–4 models) modestly reduce variance but do not substantially increase the average agreement.

Although the gains in average agreement are modest, we retain the ensemble SHOE score as it may reduce the influence of model-specific biases. Our user study did not show strong trends of any of the included models but is limited to 500 pairs of HOIs. For datasets that focus on more specialized domains, individual model biases are more likely to surface, and the ensemble may provide a more balanced estimate. Overall, the ablation shows that SHOE is largely insensitive to the exact choice of LLMs. As long as models of comparable capability are included, the resulting similarity scores remain stable. As LLMs evolve, SHOE can adopt newer models without changing its characteristic behavior.


\begin{table*}[t]
\centering
\small
\renewcommand{\arraystretch}{1.3}
\begin{tabular}{llccc}
\toprule
  \makecell{\textbf{Method}\\\textbf{Type}} & \textbf{Model}  
  & \makecell{\textbf{GT Miss Rate} \\ \textbf{(FN \%)}} 
  & \makecell{\textbf{Prediction Miss Rate} \\ \textbf{(FP \%)}} 
  & \textbf{SHOE mF1} \\
\midrule

\multirow{4}{*}{\textbf{Default}} 
  & LAIN~\cite{kim2025locality} (ViT-B)        & 21.04 & 37.77 & 45.80 \\
  & CMMP~\cite{lei2024exploring} (ViT-L)        & 20.06 & 37.64 & 50.47 \\
  & ADA-CM~\cite{lei2023efficient} (ViT-L)      & 19.18 & 36.21 & 51.93 \\
  & HOLA~\cite{lei2025hola} (ViT-L)        & 19.45 & 35.83 & 52.21\\
\midrule

\multirow{4}{*}{\makecell{\textbf{Zero-shot}\\ \textbf{RF-UC}}} 
  & CMMP~\cite{lei2024exploring} (ViT-B)        & 21.15 & 39.13 &   44.84 \\
  & EZ-HOI~\cite{lei2024EZ_HOI} (ViT-B)      & 20.45 & 38.10 &   46.94\\
  & HOLA~\cite{lei2025hola} (ViT-B)        & 20.00 & 37.29 &   49.07\\
  & LAIN~\cite{kim2025locality} (ViT-B)        & 20.83 & 38.20 &   46.30 \\
\midrule

\multirow{2}{*}{\makecell{\textbf{Open-}\\\textbf{Vocabulary}}} 
  & CMD-SE~\cite{lei2024CMD_SE}              & 36.93 & 60.67 & 37.07 \\
  & THID~\cite{wang2022learning}                & 47.81 & 78.86 & 28.71 \\
  
\midrule

\multirow{5}{*}{\makecell{\textbf{VLMs \&}\\\textbf{MLLMs}}} 
  & Qwen2.5-32B-VL~\cite{bai2025qwen25VL}   & 30.48 & 60.75 & 54.40 \\
  & InternVL3-38B~\cite{chen2024internvl}    & 55.09 & 86.10 & 40.04 \\
  & LLaVa-Onevision-72B~\cite{li2024llavaonevisioneasyvisualtask}  & 32.04 & 64.13 & 54.50 \\
  & GPT-4.1~\cite{2025Achiamgpt}          & 27.86 & 55.10 & 53.46 \\
  & Gemini-2.5-flash~\cite{comanici2025gemini} & 34.39 & 59.05 & 51.16  \\
  
\bottomrule
\end{tabular}
\caption{Performance comparison across different HOI models on the single HOI HICO-DET~\cite{chao2018learning} subset (575 classes) using GT Miss Rate, Prediction Miss Rate and SHOE mF1 scores on HICO-DET, along with GT miss rate (false negatives) and prediction miss rate (false positives). confidence score rank $i=0.5$ was chosen for all models that provide confidence scores to make them comparable to the VLM and MLLM methods.}
\label{tab:hoi_results_full}
\end{table*}


\section{Open-Vocabulary Evaluation of \\ HOI Models}

We compare Vision-Language Models (VLMs) with state-of-the-art HOI methods in Table \ref{tab:hoi_results_full}. As VLMs cannot reliably provide detection bounding boxes or confidence scores for HOI class prediction, we use the off-the-shelf object detection, DETR~\cite{carion2020end}, commonly adopted in two-stage HOI detection methods, and VLMs' token probabilities for mAP-style ranked evaluation.
However, not all VLMs provide token probabilities. 
Additionally, using DETR as an initial object detector omits the VLMs' capabilities to detect human, object and interaction, as it only has to find the interaction for the already predefined human-object pair. 
To extend our study, we report results for the confidence-free evaluation mode (i.e., category (2) confidence-free mode, see Sec. 3.2 in the main paper) for all models.
We add Gemini-2.5-flash \cite{comanici2025gemini} and LLaVa-Onevision-72b \cite{liu2024LLaVA} in this comparison as token-wise probabilities are not needed for this evaluation setting.
%
We present the GT miss rate (false negatives), prediction miss rate (false positives), and SHOE mF1 for all the models included in our evaluation. On the single-HOI HICO-DET~\cite{chao2018learning} subset (see Table \ref{tab:hoi_results_full}). We define: 

\begin{itemize}
    \item \textbf{GT Miss Rate (FN \%)}: The percentage of ground-truth HOI instances that are not matched by any prediction (false negatives):
    \[
    \text{GT Miss Rate} = \frac{|\text{FN}|}{|\text{GT}|} \times 100
    \]
    where $|\text{GT}|$ is the number of ground-truth HOI instances, and $|\text{FN}|$ is the number of ground-truth instances with no matched prediction (e.g., under the chosen IoU and similarity threshold).
    
    \item \textbf{Prediction Miss Rate (FP \%)}: The percentage of predicted HOI instances that do not match any ground-truth instance (false positives):
    \[
    \text{Prediction Miss Rate} = \frac{|\text{FP}|}{|\text{Pred}|} \times 100
    \]
    where $|\text{Pred}|$ is the total number of predictions, and $|\text{FP}|$ is the number of predictions with no matched ground-truth.
\end{itemize}
For models that provided confidence scores with their predictions, we set the confidence score rank to $i=0.5$.
We report that HOLA~\cite{lei2025hola} (ViT-L) achieves the highest SHOE mF1 score (52.21) among existing fully supervised methods, while maintaining low miss rates for both ground-truth and predictions. 
HOLA (ViT-B) performs best in the zero-shot RF-UC setting, with a SHOE mF1 score of 49.07. 
Open-vocabulary models such as CMD-SE~\cite{lei2024CMD_SE} and THID~\cite{wang2022learning} have higher miss rates and lower mF1 scores, highlighting the challenge of open-world generalization. 
These findings are consistent with the model ratings observed on the full HICO-DET dataset.
Notably, the VLMs and MLLMs perform comparably to or even outperform the existing HOI detection methods, despite lacking HOI-specific supervision, and exhibiting high GT and prediction miss rates. 
This is consistent with our observation that these models tend to make fewer predictions per ground-truth instance than standard HOI methods. 
The strong overall performance suggests that VLMs and MLLMs produce semantically plausible interaction predictions and generalize interaction concepts via language grounding.

\section{Hardware Requirements and Runtime}
In the following we provide an overview of the hardware used and the runtime required to compute the object and verb SHOE-similarity matrices on HICO-DET. These matrices have to be computed only once per dataset which enables all subsequent experiments to reuse them without any additional model inference. This section is meant as a practical reference for researchers designing new datasets, illustrating the one-time computational cost required to generate verb–verb and object–object similarity matrices.

Our setup combined an on-premise server with 4× NVIDIA RTX 6000 Ada (48 GB) GPUs, cloud inference via the Lambda service for selected models, and the Gemini-2.5-Pro API.
We first evaluated the full set of \(\approx\)850k verb–verb pairs with Qwen3-32B, which required \(\approx\)6 days on the RTX 6000 Ada system. From these, we extracted the \(\approx\)120k non-zero similarity pairs and ran them on DeepSeek-V3-0324, Llama-4-Maverick-17B, and Yi-1.5-34B-Chat, each completing in \(\approx\)30 h. For the 40k object–object pairs, all models required \(\approx\)6 h each. DeepSeek and Llama were executed on Lambda cloud instances with runtimes matching the on-premise hardware, while Gemini-2.5-Pro completed both verb and object evaluations under batched API requests within \(\approx\) 28h. To reduce wall-clock time, several runs were parallelized across cloud machines. The total cost of API usage and cloud inference was \(\approx\) 200\$ for the full HICO-DET similarity evaluation.

\begin{figure}[t]
    \centering
    \includegraphics[width=0.9\linewidth]{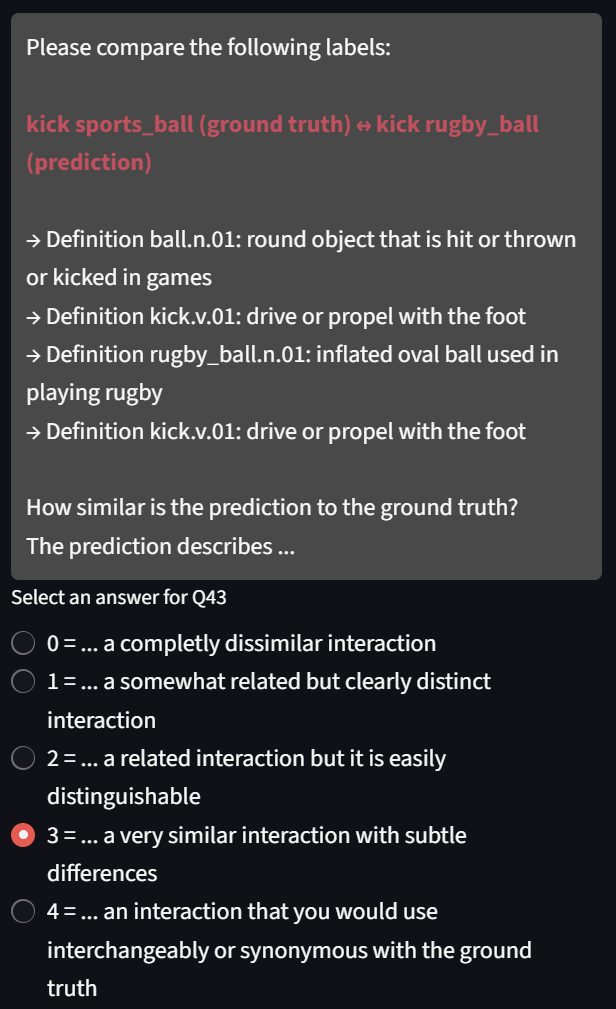}
    \caption{Example of the user study interface. Annotators are shown two HOI interactions along with their WordNet glosses and asked to rate their semantic similarity on a 5-point scale.}
    \label{fig:user_study_example}
\end{figure}

\section{Synset Matching Procedure}
Open-vocabulary predictions must first be mapped to their corresponding WordNet~\cite{wordnet} synsets before they can be evaluated with the SHOE metric. This requires converting the model outputs into consistent lexical forms before matching them to WordNet. Predicted verbs that are not in base form are lemmatized (e.g., "riding" to "ride"), and object names containing underscores are split into their separate words. 
Each normalized lemma is then queried in WordNet to retrieve all associated synsets (i.e., possible senses). 
For multi-word objects such as ``dining\_table", we query each component word separately and aggregate the valid synsets as a candidate pool. 
During evaluation, the synset from this pool with the highest semantic alignment to the ground-truth synset is selected, and this best-matching sense is used for the SHOE similarity computation.

\begin{figure}[t]
    \centering
    \includegraphics[width=0.95\linewidth]{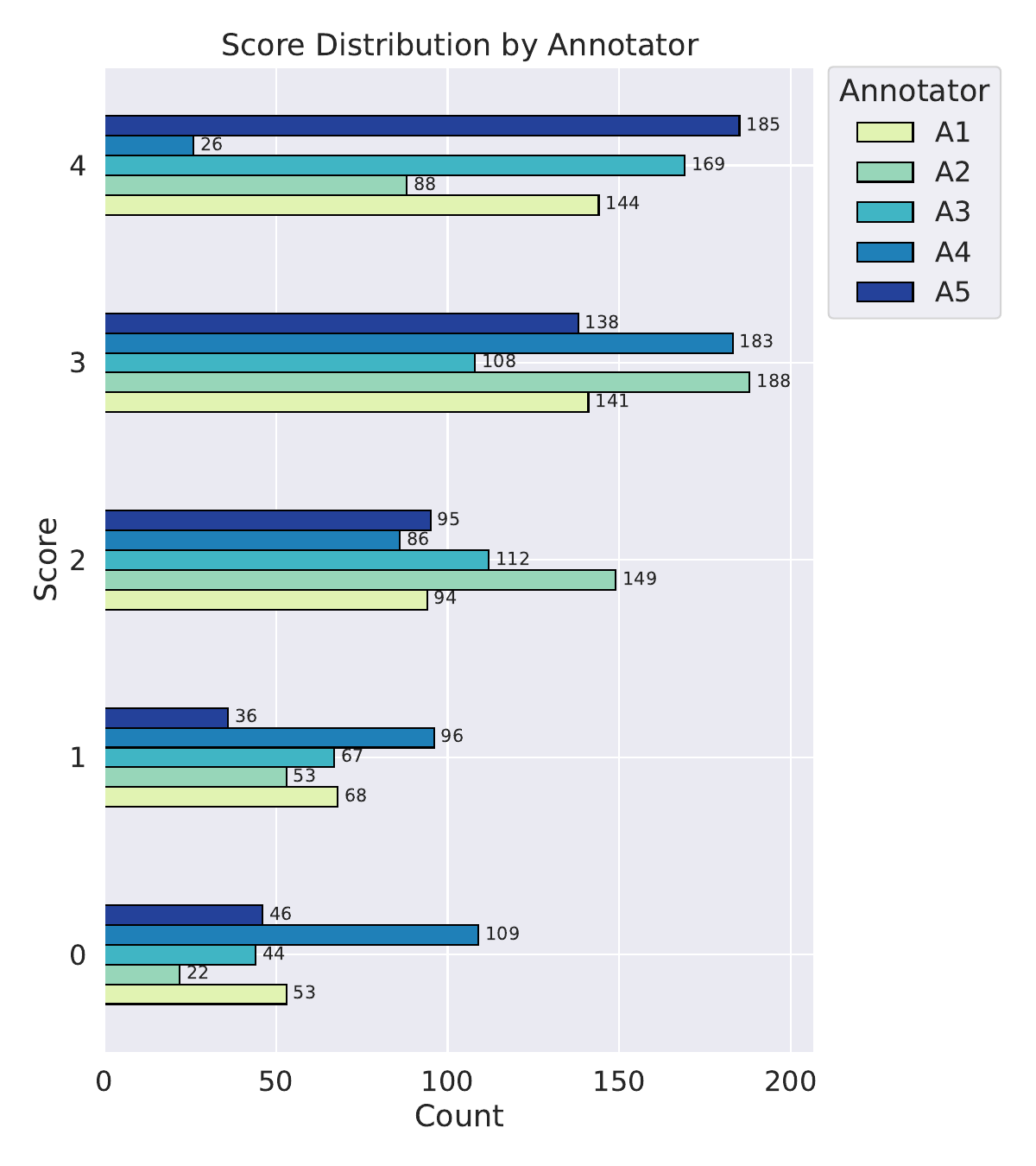}
    \caption{
        Per-annotator distribution of HOI similarity scores ranging from 0 (no similarity) to 4 (high similarity), showing individual rating tendencies across the annotation set.
    }
    \label{fig:distribution}  
\end{figure}

\section{User Study}

\subsection{User Study Setup and Examples}

To better illustrate the design of our user study, we provide a representative example of the annotation interface and rating process.
In each task, annotators were presented with a pair of HOI labels, along with the corresponding WordNet synsets and their glosses (definitions) for both the verb and object components. 
Annotators were asked to assess the semantic similarity between the two HOIs based on the information provided.
Figure~\ref{fig:user_study_example} shows an example of the annotation interface used in our study. 
The two HOIs are displayed along with their WordNet glosses to help annotators understand the meaning of each interaction. Annotators then assign a similarity score on a 5-point scale ranging from 0 (completely dissimilar) to 4 (interchangeable).

\subsection{Inter-Annotator Score-Distribution}

We analyzed the overall score distribution across annotators to examine how consistently they used the 0–4 rating scale (Figure \ref{fig:distribution}). 
All annotators used the full range of scores, though each showed distinct rating tendencies.
For example, A4 assigned score 0 more frequently than others, while A5 used score 4 most often.
A2 and A3 tended to assign scores clustered around the middle of the scale. 
Although the evaluation criteria were consistent across annotators, their scoring patterns suggest varying degrees of strictness.

\begin{figure}[t]
    \centering
    \includegraphics[width=0.9\linewidth]{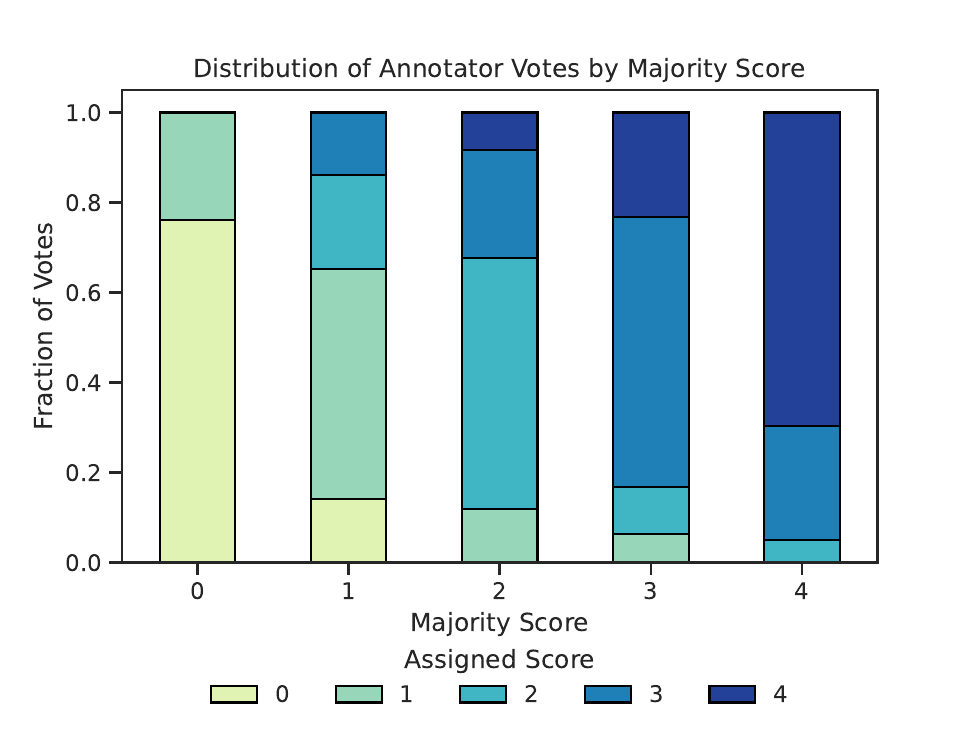}
    \caption{
        Distribution of Annotator Votes by Majority Scores. Each bar shows the number of times a given annotator (A1–A5) assigned each score from 0 to 4. The numeric labels on top of each bar indicate the raw count.
    }
    \label{fig:majorityscorevote}
\end{figure}

To better understand how annotators varied in their assignments depending on the overall consensus of an item, we grouped examples by their majority (most frequent) score and visualized the distribution of all annotator votes for each group (Figure \ref{fig:majorityscorevote}). 
Items with a majority score of 0 or 4 showed strong agreement, with most annotators assigning the same score. 
In contrast, examples with mid-range majority scores (e.g., 2) showed more varied responses across neighboring scores, reflecting greater subjectivity in these cases.

\subsection{Inter-Annotator Agreement Metrics}

\begin{figure}[t]
    \centering
    \includegraphics[width=0.85\linewidth]{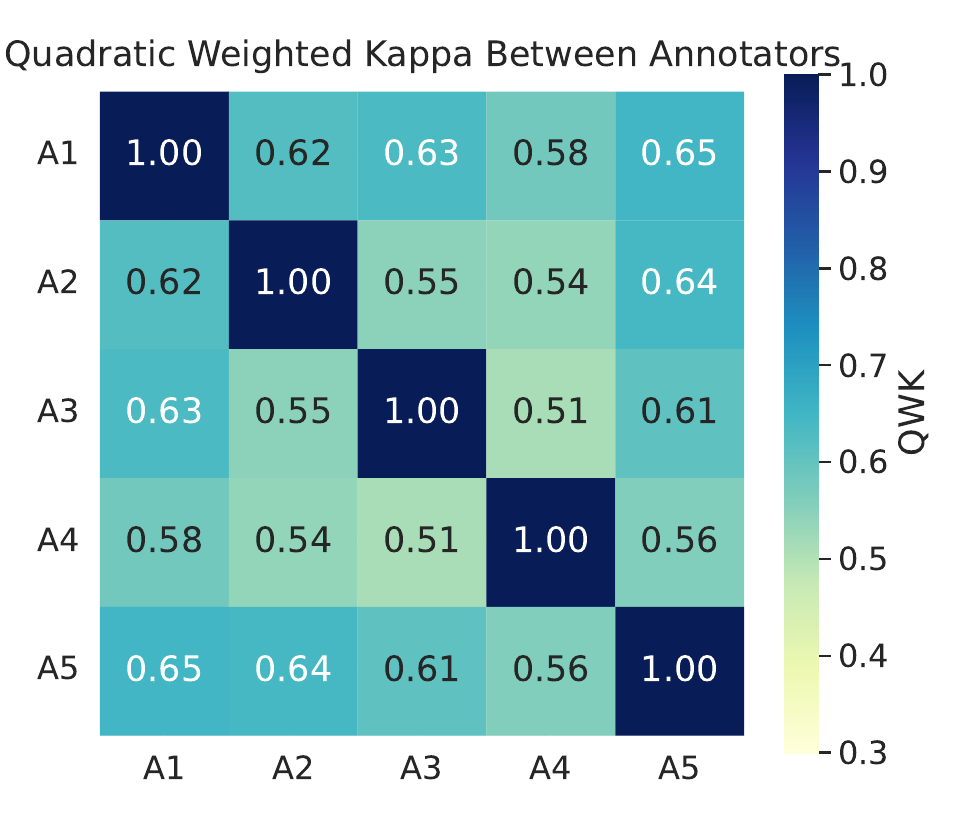}
    \caption{
        Pairwise Quadratic Weighted Kappa (QWK) between annotators.
        All values are between 0.51 and 0.65, indicating moderate agreement.
        Diagonal entries represent perfect self-agreement (\( \kappa = 1.00 \)).
    }
    \label{fig:qwk_heatmap}
\end{figure}

For the assessment of inter-annotator reliability we employed Quadratic Weighted Kappa (QWK) and Krippendorff’s alpha that account for the ordinal structure of the data: 

\paragraph{Quadratic Weighted Kappa.}
Quadratic Weighted Kappa (QWK)~\cite{fleiss1973equivalence} measures agreement between two annotators, assigning partial credit for near matches. It is defined as:

\[
\kappa = 1 - \frac{\sum_{i,j} w_{ij} \, O_{ij}}{\sum_{i,j} w_{ij} \, E_{ij}},
\]

where \( O_{ij} \) is the observed count of examples where rater A assigned score \( i \) and rater B assigned score \( j \), and \( E_{ij} \) is the expected count of such pairings under random chance. The weight matrix \( w_{ij} \) penalizes disagreements quadratically:

\[
w_{ij} = \frac{(i - j)^2}{(k - 1)^2},
\]

with \( k = 5 \) being the number of discrete rating levels. Figure \ref{fig:qwk_heatmap} shows a moderate level of agreement overall, with stronger consistency between pairs such as A1 \& A5 and A2 \& A5, and slightly weaker alignment between A3 \& A4.
The average QWK score across all annotator pairs was 0.590, indicating moderate agreement.

\paragraph{Krippendorff's Alpha (Ordinal).}
Krippendorff’s Alpha~\cite{krippendorff2018content} generalizes agreement measurement to multiple annotators and missing values. For ordinal labels, it penalizes disagreement based on the normalized distance between scores:

\[
\alpha = 1 - \frac{D_o}{D_e},
\]

where \( D_o \) is the observed disagreement, and \( D_e \) is the expected disagreement under chance. The pairwise disagreement between scores \( i \) and \( j \) is calculated as:

\[
\delta(i, j)^2 = \left( \frac{i - j}{k - 1} \right)^2.
\]

The Krippendorff’s alpha for ordinal ratings was 0.566, supporting the finding of moderate consistency among annotators. 
These results suggest that while annotators may assign slightly different scores, they generally agree on the relative severity or quality of the examples. 
This indicates that the annotation process is reliable, as the ordinal structure of the labels is largely preserved across annotators.

\section{SHOE-Metric}
\subsection{Verb and Object based LLM agreement with human annotators}
The two settings in Table~\ref{tab:userstudyagreement} show that all LLMs achieve strong agreement with human annotators if either object or verb is changed while the other is the same as the GT. Object mismatches show slightly higher agreement over all LLMs than verb mismatches. DeepSeek-V3-0324 shows the highest agreement in both categories. Our SHOE score, which aggregates model predictions through majority voting, demonstrates competitive agreement in both settings and reliability for different types of prediction mismatches.

\subsection{User Study with Images}
Our primary user study was conducted without images in order to collect semantic similarity judgments that generalize across different scenes, not tied to any specific visual instance. Including images risks biasing annotators toward incidental scene details, such as lighting, occlusion, or unusual body poses, that have nothing to do with the semantic relationship between the interactions and are visual noise that HOI models are meant to filter out. 
However, as SHOE is ultimately applied to image-conditioned HOI predictions, we conducted a second user study to verify that semantic judgments remain stable when images are actually present. 
The annotators evaluated 500 HOI pairs, each accompanied by an image from the HICO-DET dataset corresponding to the ground-truth HOI class in the pair.
They were tasked to decide if they wanted to update their original annotation or keep it the same with the image present. Across all 500 HOI pairs, only 7.65\% of ratings were updated (95\% CI: 6.29–9.00\%). The magnitude of these changes was small, only a single point on the 0-4 point scale (\(\approx\)1.14 [95\% CI 1.11, 1.17]). This demonstrates that the semantic judgments collected without images remain consistent even when annotators are later presented with visual context further supporting the reliability of our metric.

\subsection{Statistical Basis for Participant and Annotation Counts}
Our user study evaluates 500 HOI pairs through ratings from five independent annotators. To verify that this sample size is statistically adequate, we compute the Spearman correlation between model scores and human ratings. We obtain $\rho$ = 0.799 with a 95\% Fisher confidence interval (CI) of [0.765, 0.829]. This interval represents the range of plausible values for the true correlation if the study were repeated many times with the same sample size. The interval is narrow and strongly positive. This supports that 500 items provide a robust estimate of the true correlation. We also performed a permutation test with 10,000 resamples, which tests if such a high correlation could occur under random ratings. The result $p < 0.001$ indicates that the observed correlation is extremely unlikely to arise by chance.

\begin{table}[t]
\centering
\small
\renewcommand{\arraystretch}{1.4}
\resizebox{\linewidth}{!}{%
\begin{tabular}{p{0.4\linewidth} >{\centering\arraybackslash}p{0.25\linewidth} >{\centering\arraybackslash}p{0.25\linewidth}}
\toprule
\textbf{Metric Model} &
\textbf{Verb Agreement (\%)} &
\textbf{Object Agreement (\%)} \\
\midrule
Standard SHOE        & 78.82 & 80.09 \\
DeepSeek-V3-0324     & 79.89 & 81.21 \\
Qwen3-32B            & 78.86 & 81.41 \\
Yi-1.5-34B-Chat      & 79.01 & 80.61 \\
Llama-4-Maverick-17B & 77.92 & 80.99 \\
Gemini-2.5-pro       & 78.34 & 81.34 \\
\bottomrule
\end{tabular}%
}
\caption{Average LLM agreement with human annotator ratings on HOI pairs where either the verb or object differs from the ground truth while the other is identical.}
\label{tab:userstudyagreement}
\end{table}

To assess whether five annotators per item are sufficient, we estimate the within-item variability ($\sigma$ \(\approx\) 0.761) and compute the standard error of the mean rating ($\mathrm{SE} = \frac{\sigma}{\sqrt{k}}$).
Five raters achieved SE \(\approx\) 0.34. This SE implies that the mean rating would vary by 0.34 of a point across different groups of annotators, which is small relative to the full rating range (0-4) and sufficient to distinguish meaningful differences between items. This shows that five annotators are enough to provide reliable similarity scores.

\subsection{SHOE-algorithms}
To make our work more reproducible we describe the implementation of the two SHOE-scoring algorithms.
Our main SHOE-mAP-score algorithm is shown in Algorithm \ref{alg:map}. For each image, predictions are filtered by a confidence threshold and matched to ground-truth instances based on the common IoU $>$ 0.5 requirement on the human and object bounding boxes. Predictions count as soft true positives when they match spatially and have highest semantic similarity of all predictions for that GT. Predictions that do not match any ground truth get penalized through false positive penalties. For each unmatched prediction, the GT class with the highest similarity score is found. A FP is assigned to that class if the similarity score is larger than a certain threshold. The algorithm sorts predictions by confidence before it calculates true positives and false positives using soft labels for each class. Precision and recall get computed at each threshold before obtaining Average Precision (AP) through the calculation of area under the resulting precision-recall curve. The final SHOE mAP is the mean of AP scores across all classes.

The procedure for computing the SHOE mF1 score for HOI predictions using soft matching based on semantic similarity is described in Algorithm \ref{alg:mf1}. If models provide confidence scores only predictions larger than a chosen threshold are included in the calculation ($s \geq \tau$). For each image the spatial alignment of each prediction is verified through bounding box IoU and the semantic similarity between predicted verb-object pairs and ground truth pairs is computed with our provided SHOE similarity maps. Predictions get matched identically to the SHOE-mAP-score. In the final aggregation step precision, recall and F1 scores for each class are calculated based on the accumulated FP, TP and FN before averaging over all classes for the mean F1 score (mF1).

\section{Future Work}
While large VLMs show strong semantic reasoning for HOI prediction, they currently rely on an external detector such as DETR due to their limited detection capabilities. VLMs require substantially more parameters (\(\approx\)1.5B - \(\approx\)400B) compared to standard HOI detectors (\(\approx\)40M), making them impractical for real-time deployment and HOI detection from videos. Their prompt sensitivity and non-deterministic outputs further complicate consistent predictions and integration into downstream systems. This highlights the need to develop true open-vocabulary, real-time HOI predictors that combine strong localization with true semantic understanding.

On the dataset side, existing HOI benchmarks often focus on relatively small subsets of interactions and do not exhaustively annotate all visible HOIs within each image often using broad action categories like "cooking", "working", or "cleaning" but miss lower-level HOIs like "cutting", "holding", "brushing". As the field moves toward open-vocabulary HOI recognition, it will be essential to establish clear annotation guidelines and to develop more comprehensive datasets.


\begin{figure*}[t]
\centering

\begin{minipage}[t]{0.47\textwidth}
\textbf{Algorithm 1: SHOE mAP Evaluation for HOI Predictions}
\vspace{0.5em}
\begin{algorithmic}[1]
\label{alg:map}
\State Load ground truth set $\mathcal{G} = \{(v, o, b_h, b_o)\}$
\State Load predictions $p = (v', o', b'_h, b'_o, s) \in \mathcal{P}$
\State Load similarity maps $S_v$, $S_o$
\State Initialize per-class lists $\mathcal{S}_k$, $\mathcal{L}_k$, and GT counts $\mathcal{N}_{\text{GT}}[k] = 0$

\For{each image $I$}
    \State $\mathcal{G}_I \gets$ ground truth for $I$
    \State $\mathcal{P}_I \gets$ predictions for $I$
    \State Initialize matched flags for $\mathcal{P}_I$

    \For{each $(v, o, b_h, b_o) \in \mathcal{G}_I$}
        \State $k \gets$ class ID of $(v, o)$
        \State $\mathcal{N}_{\text{GT}}[k] \mathrel{+}= 1$
        \State $\mathcal{M} \gets \{$ predictions in $\mathcal{P}_I$ where
        \Statex \hspace{2em} $\operatorname{IoU}(b_h, b'_h) \geq \theta$ and $\operatorname{IoU}(b_o, b'_o) \geq \theta$
        \Statex \hspace{2em} and prediction not already matched $\}$
        \If{$\mathcal{M} \neq \emptyset$}
            \State Select $p^* = \arg\max\limits_{p \in \mathcal{M}}$
            \Statex \hspace{5.2em} $\frac{1}{2}(S_v(v, v') + S_o(o, o'))$
            \State Let $p^* = (v'^*, o'^*, b'^*_h, b'^*_o, s^*)$, 
            \Statex \hspace{5.2em}where $p^* \in \mathcal{M} \subseteq \mathcal{P}$
            \State Compute similarity score 
            \Statex \hspace{5.2em} $\sigma = \frac{1}{2}(S_v(v, v'^*) + S_o(o, o'^*))$
            \State Append confidence and similarity score 
            \Statex \hspace{5.2em} $(s^*, \sigma)$ to $\mathcal{S}_k, \mathcal{L}_k$
            \State Mark $p^*$ as matched
        \Else
            \State Append empty pair $(0.0, 0.0)$ to $\mathcal{S}_k, \mathcal{L}_k$
        \EndIf
    \EndFor

     \For{each unmatched $p = (v', o', b'_h, b'_o, s) \in \mathcal{P}_I$}
        \State Find $(v^*, o^*) \in \mathcal{G}_I$ maximizing similarity
        \State $\sigma = \frac{1}{2}(S_v(v^*, v') + S_o(o^*, o'))$
        \If{$\sigma \geq \delta$}
            \State $k^* \gets$ class ID of $(v^*, o^*)$
            \State Append $(s, 0.0)$ to $\mathcal{S}_{k^*}, \mathcal{L}_{k^*}$ 
        \EndIf
    \EndFor
\EndFor

\For{each class $k = 1 \dots K$}
    \State Sort $\mathcal{S}_k, \mathcal{L}_k$ by descending score
    \State Let $\epsilon$ be a small constant 
    \Statex \hspace{1em} (e.g., $10^{-8}$) to avoid division by zero
    \State Compute cumulative true positives:
    \Statex \hspace{1em} $\text{TP}_i = \sum_{j=1}^i \mathcal{L}_k[j]$
    \State Compute cumulative false positives: $\text{FP}_i = i - \text{TP}_i$
    \State $\text{Prec}_i = \text{TP}_i / (\text{TP}_i + \text{FP}_i + \epsilon)$ 
    \State $\text{Rec}_i = \text{TP}_i / (\mathcal{N}_{\text{GT}}[k] + \epsilon)$
    \State Compute $\text{AP}_k$ as area under the PR-curve
\EndFor

\State Report $\text{mAP} = \frac{1}{K} \sum_{k=1}^K \text{AP}_k$
\end{algorithmic}
\end{minipage}
\hfill
\begin{minipage}[t]{0.47\textwidth}
\textbf{Algorithm 2: SHOE mF1 Evaluation for HOI Predictions}
\vspace{0.5em}
\begin{algorithmic}[1]
\label{alg:mf1}
\State Load ground truth set $\mathcal{G} = \{(v, o, b_h, b_o)\}$
\State Load predictions $p = (v', o', b'_h, b'_o, s) \in \mathcal{P}$ 
\Statex with confidence score rank $s \geq \tau$
\State Load similarity maps $S_v$, $S_o$
\State Initialize TP = FP = FN = 0 per class

\For{each image $I$}
    \State $\mathcal{G}_I \gets$ ground truth for $I$
    \State $\mathcal{P}_I \gets$ predictions for $I$
    \State Initialize matched flags for $\mathcal{P}_I$

    \For{each $(v, o, b_h, b_o) \in \mathcal{G}_I$}
        \State $\mathcal{M} \gets \{$ predictions in $\mathcal{P}_I$ where
        \Statex \hspace{2em} $\operatorname{IoU}(b_h, b'_h) \geq \theta$ and $\operatorname{IoU}(b_o, b'_o) \geq \theta$
        \Statex \hspace{2em} and prediction not already matched $\}$
        \If{$\mathcal{M} \neq \emptyset$}
            \State Select $p^* = \arg\max\limits_{p \in \mathcal{M}}$
            \Statex \hspace{5.2em} $\frac{1}{2}(S_v(v, v') + S_o(o, o'))$
            \State Let $p^* = (v'^*, o'^*, b'^*_h, b'^*_o, s^*)$, 
            \Statex \hspace{5.2em}where $p^* \in \mathcal{M} \subseteq \mathcal{P}$
            \State Compute similarity score 
            \Statex \hspace{5.2em} $\sigma = \frac{1}{2}(S_v(v, v'^*) + S_o(o, o'^*))$
            \State Accumulate TP for class $(v, o)$ as $\sigma$
            \State Accumulate FP for class $(v, o)$ as $1-\sigma$
            \State Mark $p^*$ as matched
        \Else
            \State Accumulate FN for class $(v, o)$ as 1.0
        \EndIf
    \EndFor

    \For{each unmatched $p = (v', o', b'_h, b'_o) \in \mathcal{P}_I$}
        \State Find best $(v, o) \in \mathcal{G}_I$ maximizing similarity
        \State $\sigma = \frac{1}{2}(S_v(v, v') + S_o(o, o'))$
        \If{$\sigma \geq \delta$}
            \State Accumulate FP for class $(v, o)$ as 1.0
        \EndIf
    \EndFor
\EndFor

\For{each class $(v, o)$}
    \State Let $\epsilon$ be a small constant 
    \Statex \hspace{1em} (e.g., $10^{-8}$) to avoid division by zero
    \State Compute precision = $\frac{\text{TP}}{\text{TP} + \text{FP} + \epsilon}$
    \State Compute recall = $\frac{\text{TP}}{\text{TP} + \text{FN} + \epsilon}$
    \State Compute $F1 = \frac{2 \cdot \text{precision} \cdot \text{recall}}{\text{precision} + \text{recall} + \epsilon}$
\EndFor

\State Report mean $F1$ over all classes
\end{algorithmic}
\end{minipage}

\label{fig:shoe_algorithms}
\end{figure*}


\section{Benchmarking Datasets}
We conducted our primary evaluation on HICO-DET~\cite{chao2018learning}, as it is widely used in the HOI community. Most HOI models provide pretrained checkpoints on HICO-DET making it suitable for comparison between standard, zero-shot and open-vocabulary HOI models. SWIG-HOI~\cite{swighoi} is commonly used for open-vocabulary classification as it expands the label space to 1,000 object categories and 406 actions. However, its annotations inherit the scene-level abstraction of SWiG{~\cite{pratt2020grounded}, which causes many fine-grained HOIs to be missed or collapsed into coarse descriptions. Fig.~\ref{tab:hico_vs_swig} shows a qualitative comparison between 3 common scenarios in both HICO-DET and SWIG-HOI dataset. Specific interactions such as eat, wash, or cut are annotated within HICO-DET. However, SWIG-HOI includes broader verbs like cook or clean as their ground-truth HOI classes, because it is derived from general scene understanding. 

Moreover, SWIG-HOI includes only a single human annotation per image, even when multiple people interact with objects (see Fig.~\ref{tab:hico_vs_swig} second SWIG-HOI example "cook vegetable/knife") where two people cutting vegetables are visible in the image but only 1 person and the corresponding HOI is annotated. This is acceptable for closed-set classifiers with fixed verb–object categories but becomes problematic in open-vocabulary evaluation with no predefined categories. In such settings, a model may correctly predict a valid but unannotated human–object interaction, which is then incorrectly counted as a false positive penalizing category-free predictors. SWIG-HOI's scene-level annotation style and incomplete human coverage makes it less suitable for open-vocabulary HOI evaluation. For this reason, we chose HICO-DET for our quantitative open-vocabulary evaluation. 

V-COCO~\cite{lin2014microsoft} is an older benchmark that offers substantially less interaction categories  than the newer HICO-DET (26 actions vs 117 actions). Many HOI models do not provide checkpoints for V-COCO and focus on evaluating on HICO-DET or SWIG-HOI. To demonstrate that the SHOE metric transfers seamlessly across benchmarks we report some exemplary results on this dataset (Table \ref{tab:coco_shoe_map}). The trends of the evaluated LLMs  mirror those observed on HICO-DET. GPT-4.1 shows strongest performance for the standard mAP score while Qwen2.5-VL-32B achieves the highest SHOE mAP on V-COCO..

\begin{table}[t]
\centering
\small
\begin{tabular}{lcc}
\toprule
\textbf{Model} & \textbf{mAP} & \textbf{SHOE-mAP} \\
\midrule
GPT-4.1 & \textbf{50.47} & 57.82\\
Qwen2.5-VL-32B& 41.12 & \textbf{64.77} \\
InternVL3-38B & 46.34 & 50.79 \\
\bottomrule
\end{tabular}
\caption{
Comparison of mAP and SHOE-mAP on V-COCO for three common VLMs. VLMs token probabilities are used as a proxy for the confidence score and DETR is used for object detection.
}
\label{tab:coco_shoe_map}
\end{table}

\section{Qualitative results}
To better understand the results of Table \ref{tab:hoi_results_full}, we present qualitative examples of GPT-4.1’s predictions and their alignment with HICO-DET ground truth based on our instance-level SHOE score (see Figure \ref{fig:GPTqualiSHOE}). 
The examples show that GPT-4.1 often generates semantically plausible outputs that remain close in meaning to the intended action, even when the exact ground-truth interaction is missed.

For example, the ground-truth action “pet a giraffe” receives a high similarity score of 0.775 when matched with the prediction “touch a giraffe,” which is semantically plausible given the close relation between these actions. Similarly, “wash a train” is matched with “clean a train” with a similarity score of 0.85.
These examples highlight the ability of the VLMs and MLLMs to perform robust conceptual reasoning and demonstrate that our SHOE score can capture a broad range of semantically aligned open-vocabulary interactions.

\begin{figure*}[t]
\centering
\renewcommand{\arraystretch}{1.4}
\setlength{\tabcolsep}{12pt}

\begin{tabular}{c c}
\textbf{(a) HICO-DET} & \textbf{(b) SWIG-HOI} \\

\begin{minipage}{0.45\textwidth}
    \centering
    \includegraphics[height=4cm]{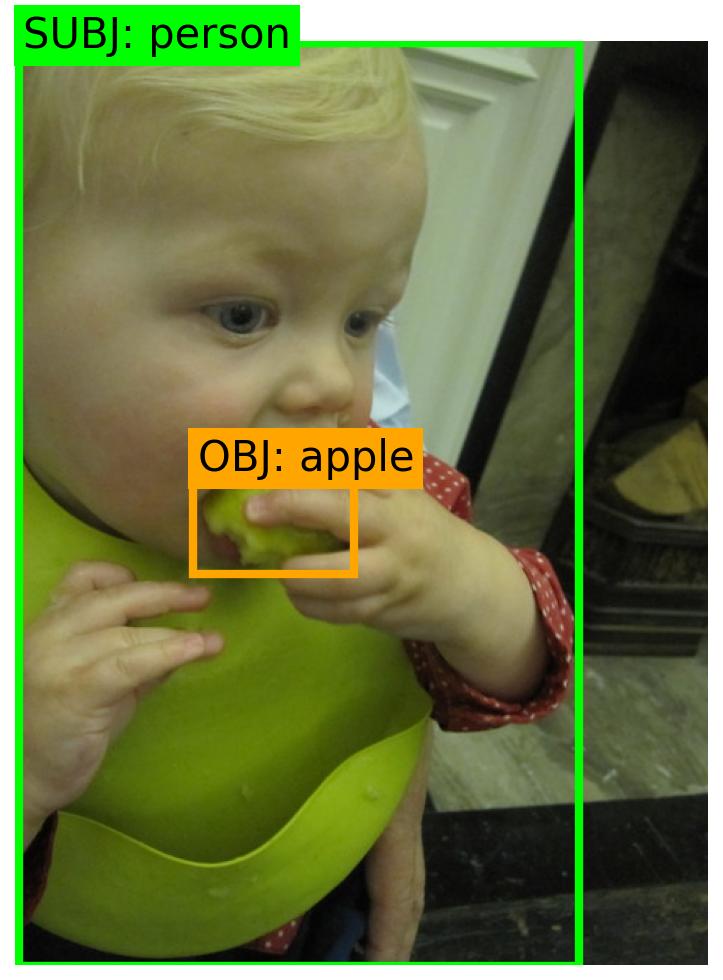}

    {\small
    \textbf{Object(s):} Apple\\
    \textbf{Interactions:} eat, hold
    }
    \vspace{4mm} 
\end{minipage}
&
\begin{minipage}{0.45\textwidth}
    \centering
    \includegraphics[height=4cm]{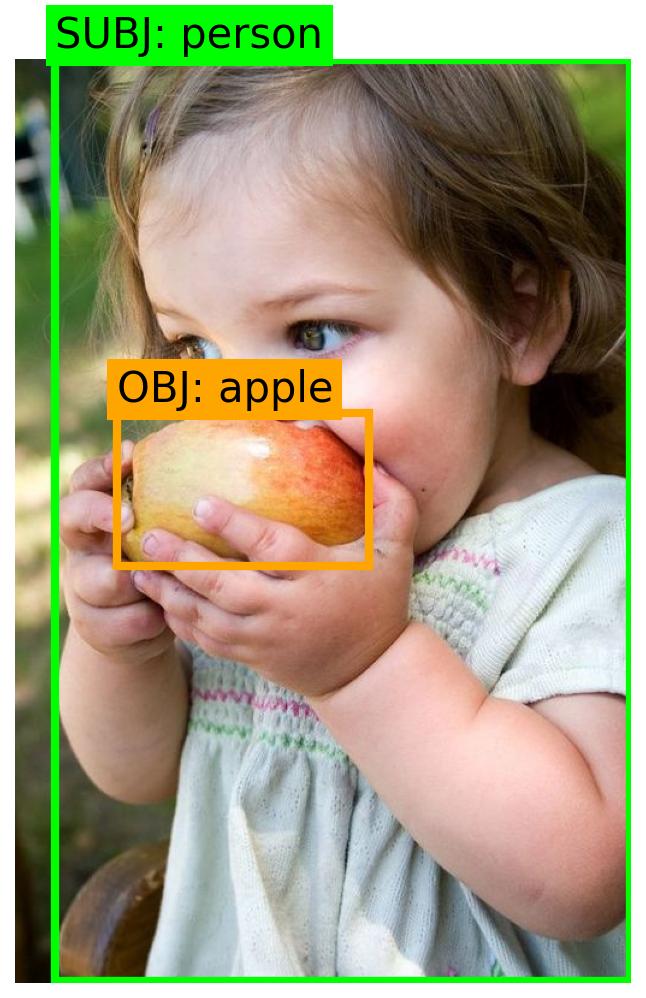}

    {\small
    \textbf{Object(s):} Apple\\
    \textbf{Interactions:} lick
    }
    \vspace{4mm} 
\end{minipage}
\\[10mm] 

\begin{minipage}{0.45\textwidth}
    \centering
    \includegraphics[height=4cm]{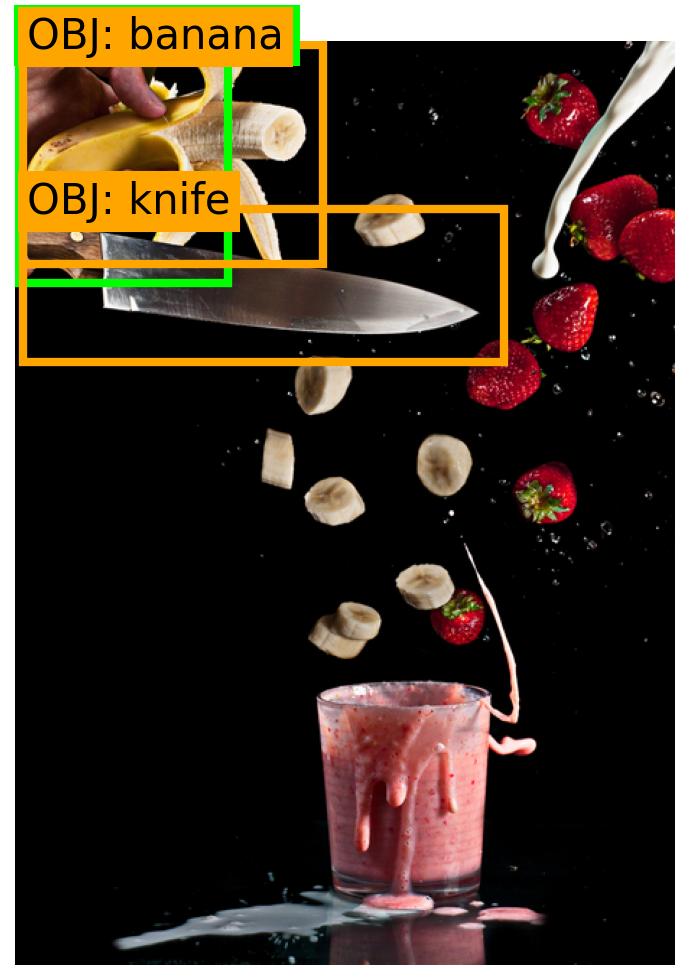}

    {\small
    \textbf{Object(s):} Banana, Knife\\
    \textbf{Interactions:} hold, cut with, wield
    }
    \vspace{4mm}
\end{minipage}
&
\begin{minipage}{0.45\textwidth}
    \centering
    \includegraphics[height=4cm]{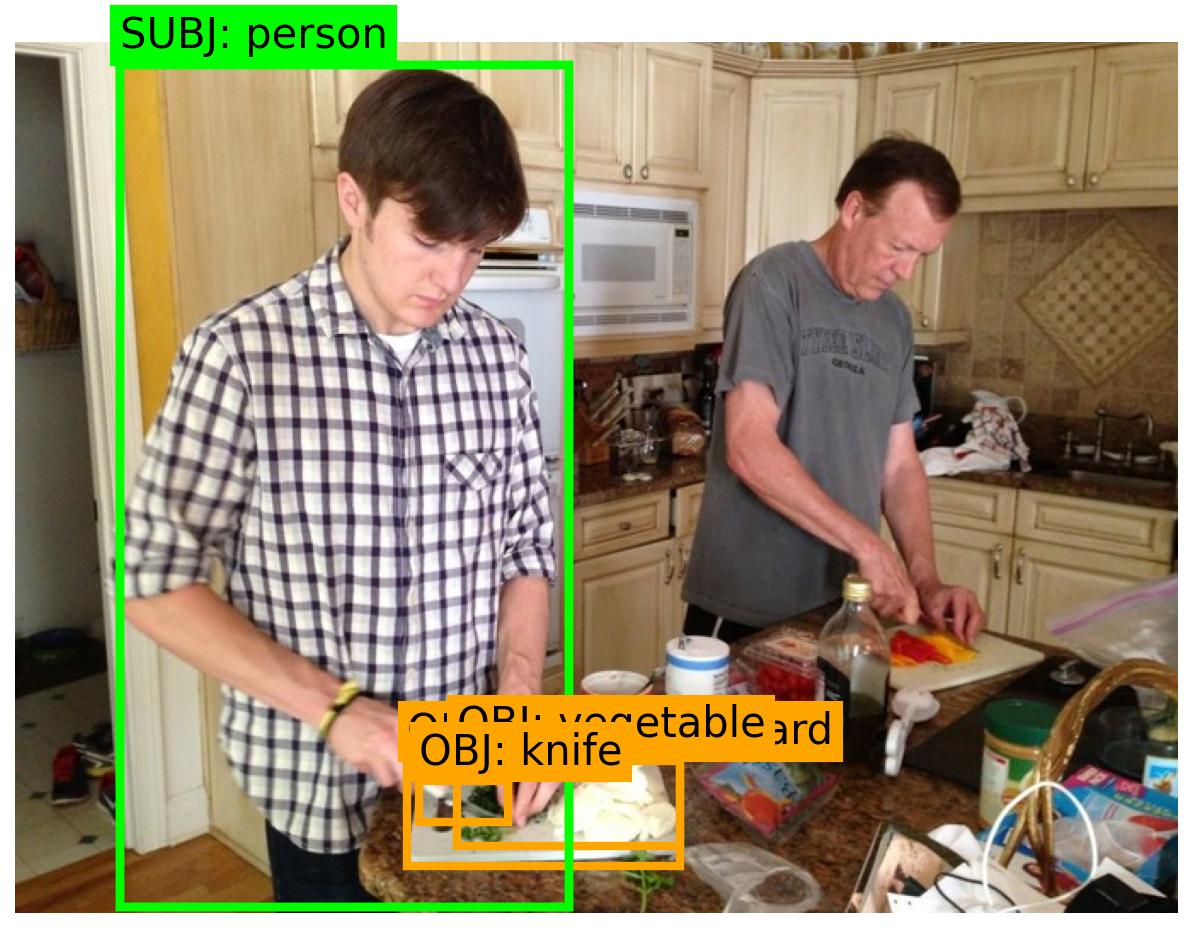}

    {\small
    \textbf{Object(s):} Knife, Vegetable\\
    \textbf{Interactions:} cook
    }
    \vspace{4mm}
\end{minipage}
\\[10mm]

\begin{minipage}{0.45\textwidth}
    \centering
    \includegraphics[height=4cm]{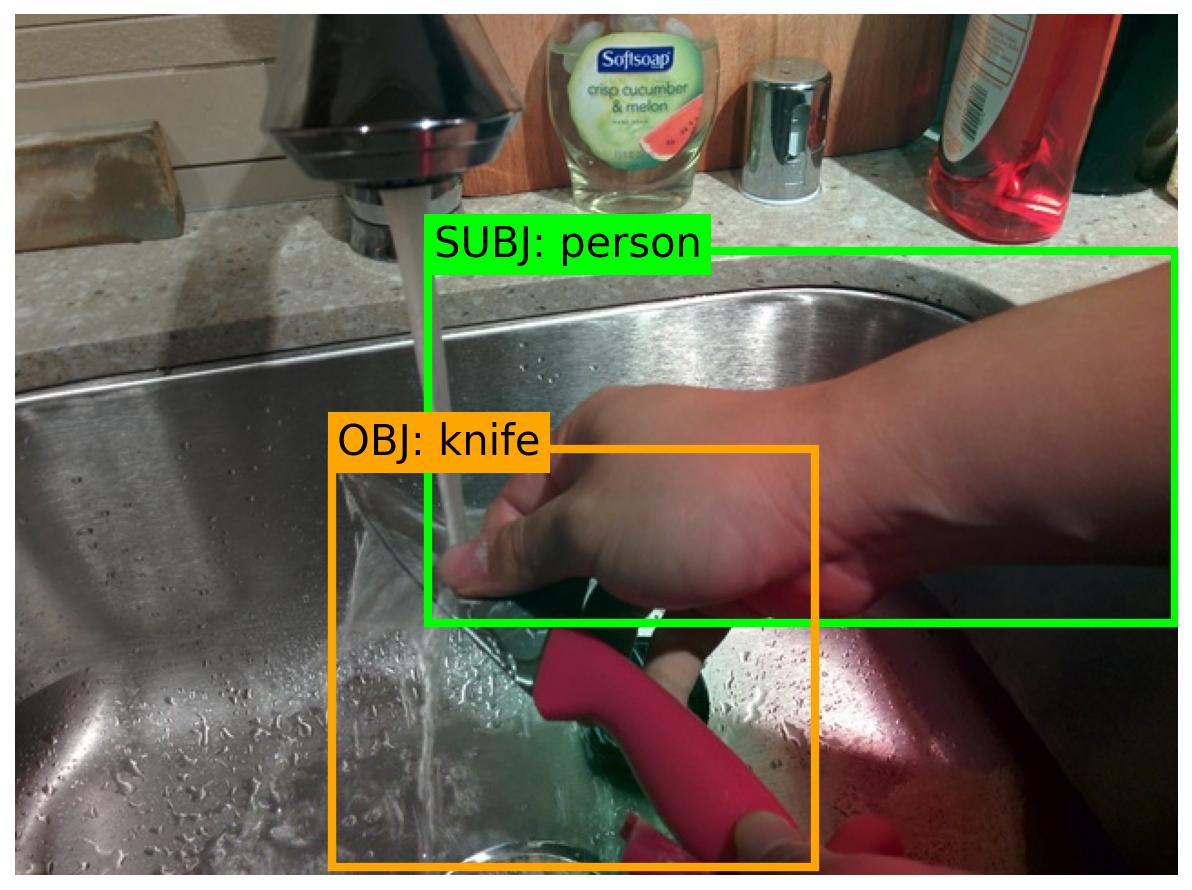}

    {\small
    \textbf{Object(s):} Knife\\
    \textbf{Interactions:} hold, wash
    }
    \vspace{2mm}
\end{minipage}
&
\begin{minipage}{0.45\textwidth}
    \centering
    \includegraphics[height=4cm]{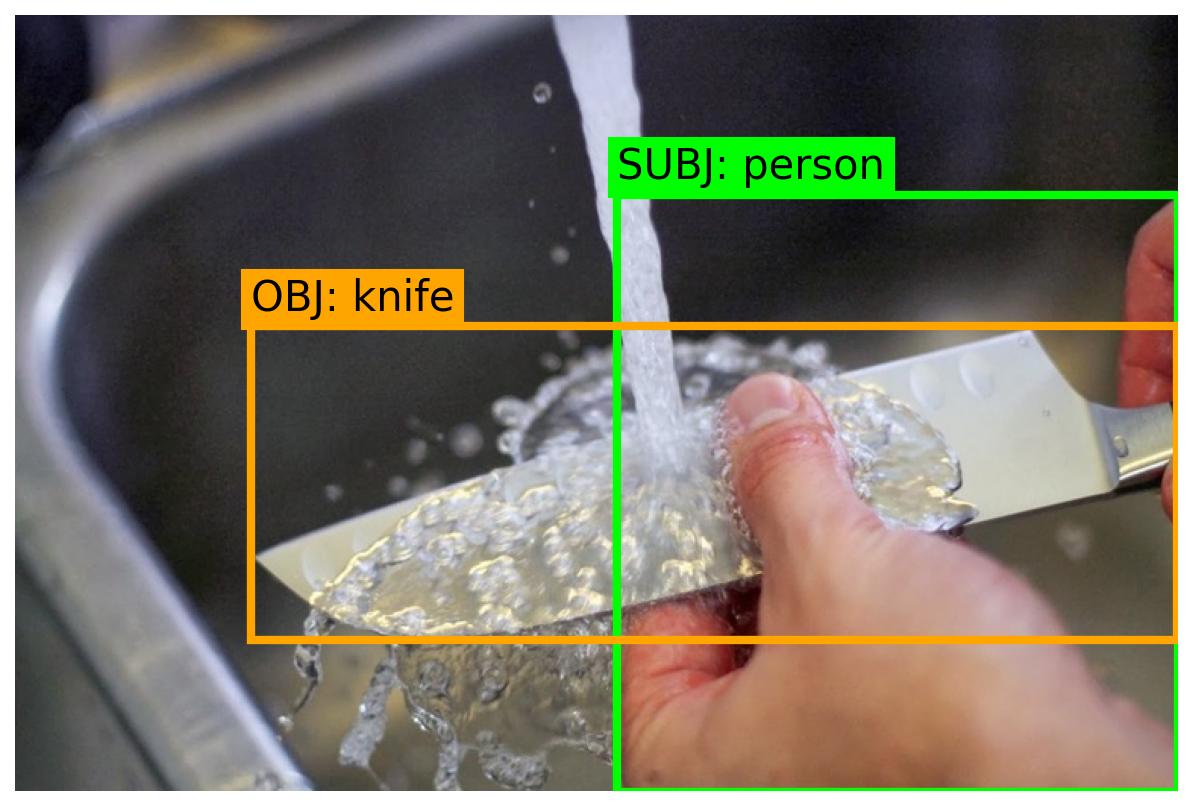}

    {\small
    \textbf{Object(s):} Knife\\
    \textbf{Interactions:} clean
    }
    \vspace{2mm}
\end{minipage}
\\

\end{tabular}

\caption{Qualitative comparison of interaction annotations. HICO-DET captures fine-grained, specific interactions (e.g., "eat", "wash", "cut with"), whereas SWIG-HOI primarily provides broad scene descriptions (e.g., "cook", "clean").}
\label{tab:hico_vs_swig}
\end{figure*}

\begin{figure*}[ht]
    \centering
    \includegraphics[width=0.85\linewidth]{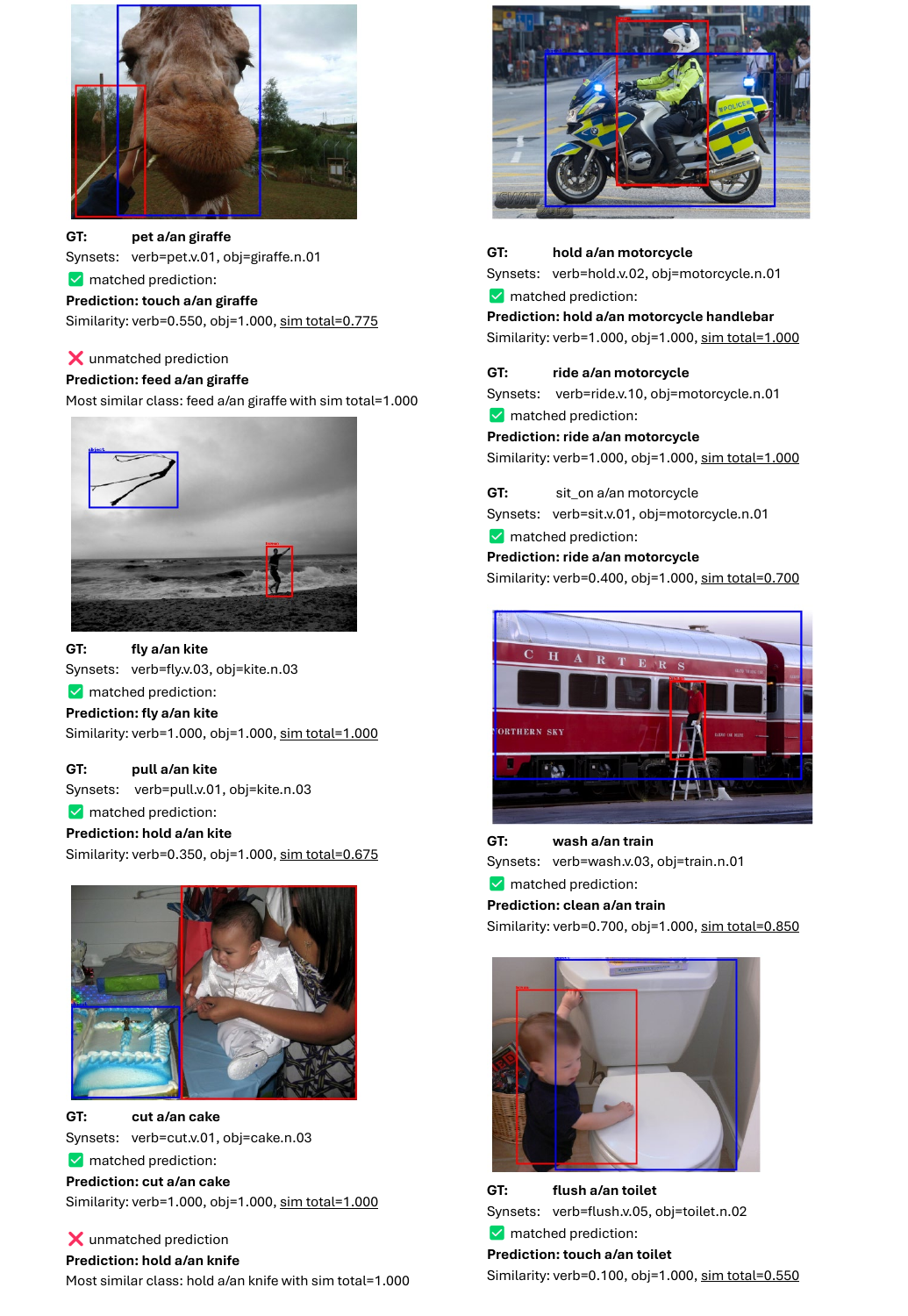}
    \caption{
        Qualitative results for GPT4.1 on HICO-DET single HOI subset.
    }
    \label{fig:GPTqualiSHOE}  
\end{figure*}

\section{Limitations}
Our proposed SHOE-Metric relies on WordNet sense disambiguation and LLM similarity scoring, both of which can introduce biases into the metric. As WordNet senses were developed by linguists primarily from English corpora it may miss coverage in domain-specific or culturally diverse verb senses. Similarly, LLM judgments can reflect biases present in their training data. As existing HOI benchmarks focus on common general interactions, we did not find any apparent biases for these datasets during testing. 
For building highly domain-specific HOI datasets with low WordNet coverage (e.g., medical or laboratory settings) we suggest using definitions from common ontologies in the field such as UMLS \cite{bodenreider2004umls} and reducing risks of domain biases by conducting an expert user study, similar to the methodology adopted in this work.